\documentclass{article}
\usepackage{hyperref}

\usepackage{multirow}
\usepackage{wrapfig}
\usepackage{subcaption}
\usepackage{amsmath}

\usepackage[main,final]{neurips_2026}

\usepackage[utf8]{inputenc} 
\usepackage[T1]{fontenc}    
\usepackage{hyperref}       
\usepackage{url}            
\usepackage{booktabs}       
\usepackage{amsfonts}       
\usepackage{nicefrac}       
\usepackage{microtype}      
\usepackage{xcolor}         
\usepackage{graphicx}
\title{\textsc{TED}: Text-Axis Evidence Decomposition for Prompted Anomaly Localization}

\author{%
JinYoung Kim$^{1}$,
Geonho Kim$^{1}$,
GiJeong Park$^{1}$,
Geonu Lee$^{2}$,
YoungJoon Yoo$^{1*}$\\
$^{1}$Department of Artificial Intelligence,
Chung-Ang University, Seoul, Korea\\
$^{2}$SNUAILAB Co., Ltd., Seoul, Korea\\
{\ttfamily\small \{barraki7226,skdmlqnsrlt,rjsgh2250,yjyoo3312\}@cau.ac.kr,}\\
{\ttfamily\small \{geonu.lee,yjyoo3312\}@snuailab.ai}
}

\begin{document}

\maketitle
{\let\thefootnote\relax\footnotetext{\textsuperscript{*}Corresponding author.}}

\begin{abstract}
CLIP is a powerful vision-language model, but it was not designed for fine-grained defect localization; CLIP-based anomaly detectors therefore adapt it with prompts or lightweight modules to increase defect sensitivity.
We show that stronger sensitivity does not necessarily make local evidence reliable: under domain shift, adapted CLIP-AD models often assign high anomaly scores to both true defects and visually complex normal regions.
The issue is not simply missing defect information, but a local scoring rule that decodes defect and hard-normal evidence, having the same anomaly evidence.
We propose \textsc{TED} (Text-Axis Evidence Decomposition), a post-hoc scoring method that asks whether each ambiguous response is better supported by source defect patches or by source normal patches mistaken as anomalous.
\textsc{TED} compares these supports under the host's normal-versus-anomaly text response, leaves the backbone and prompts unchanged, and requires no target-domain training.
It works as a train-free score for raw VLM backbones or as a source-calibrated residual correction for adapted CLIP-AD hosts.
Across frozen VLM backbones, \textsc{TED} substantially improves pixel-level localization over raw prompt similarity; across adapted hosts, it improves most pixel-level settings over P-AUROC, P-PRO, and P-AP.
Gains are largest under stronger hard-FP competition, with mean localization gain increasing from $+5.0$ in low-competition regimes to about $+10.9$ in mid/high-competition regimes.
These results suggest that recoverable defect evidence can already exist in pretrained multimodal representations, but reliable localization requires decoding it against hard-normal competitors.
Code will be released at
\href{https://github.com/barraki72268-sketch/TED-Text-Axis-Evidence-Decomposition-for-Prompted-Anomaly-Localization}{\texttt{TED GitHub repository}}.
\end{abstract}

\section{Introduction}
\label{sec:intro}

CLIP and related pretrained vision-language models (VLMs) provide a strong semantic interface for anomaly detection (AD), where normal and anomalous states can be described by language and compared with local visual features.
However, these models were not designed to localize fine-grained defect regions.
A patch can be semantically salient, structurally complex, or visually distinctive without being defective, so raw prompt similarity can assign high anomaly responses to hard normal regions.
Recent CLIP-based AD methods address this mismatch with learnable prompts, prompt ensembles, or lightweight visual modules, making the representation more defect-sensitive.
The key question is whether this stronger sensitivity yields cleaner defect evidence, or instead sharpens a local scoring rule in which true defects and visually complex normal regions remain difficult to rank apart.

Our analysis supports the latter interpretation.
Across prompt-tuned and adapter-based CLIP-AD hosts, true defects and hard false positives often compete within the same local score regime under transfer.
Normal textures, edges, reflections, and salient object parts can receive elevated anomaly responses despite being non-defective.
Thus, the failure is better viewed not as semantic loss or missing defect information, but as hard-FP entanglement: defect-supported and hard-normal-supported evidence are decoded together by raw prompt similarity or host-specific local scoring.

This entanglement makes simple false-positive suppression insufficient.
Hard false positives do not form a separable nuisance component that can be removed without affecting true defects.
Rather, the same local response that supports defect localization can also be supported by visually complex normal regions.
The relevant problem is therefore not only how to suppress false positives after they appear, but how to decide which source of evidence better explains an ambiguous local response.

To characterize this behavior, we analyze prompted VLM-based anomaly detectors through their normal-versus-anomaly text response.
Empirically, the normal-versus-anomaly text response is still useful: true defect patches usually receive stronger anomaly responses than generic normal background.
However, this response is not selective enough.
Visually complex normal regions can receive similarly high responses, so true defects and hard false positives remain difficult to rank apart.
This clarifies the role of adaptation.
Prompt tuning and lightweight modules can make the model more sensitive to defects, but they do not necessarily separate true defects from visually complex hard-normal regions in the local anomaly map.
Therefore, hard-normal structures can still be ranked close to true defects when anomaly maps are decoded by raw prompt similarity.

Motivated by this observation, we propose \textsc{TED} (Text-Axis Evidence Decomposition), a post-hoc scoring method for prompted anomaly localization.
The key idea is to not trust a high anomaly score by itself.
For each ambiguous patch, \textsc{TED} compares its response with two source banks: true defect patches and normal patches that the host previously mistook as anomalous.
The comparison is made under the same normal-versus-anomaly text response used by the host, so the query and source examples are judged in the same coordinate.
If the query is better supported by source defects, \textsc{TED} raises its score; if it is better supported by hard-normal examples, \textsc{TED} suppresses it.
Thus, \textsc{TED} separates defects from hard normal regions by asking which source evidence explains the high response.
It leaves the backbone and prompts unchanged, requires no target-domain training, and is used either as a train-free score for raw VLM backbones or as a bounded source-calibrated residual for adapted CLIP-AD hosts.

Because \textsc{TED} only re-scores local responses, it is not tightly tied to the host architecture.
Many CLIP-AD methods rely on specific CLIP components, layers, or scoring rules, so changing layers or backbones can require retraining or redesign.
By contrast, \textsc{TED} only needs patch-level visual features and a normal-versus-anomaly text response.
The same source-evidence comparison can therefore be evaluated on frozen VLM backbones while keeping the host and target protocol fixed.
Experiments across adapted CLIP-AD hosts, frozen VLM backbones, and cross-domain transfer benchmarks show that \textsc{TED} improves pixel-level localization most clearly when hard-normal evidence is ranked close to true defects.
\textsc{TED} does not simply raise all anomaly scores; it helps most when the baseline confuses hard normal regions with true defects, and less when the host already separates them well.
Overall, \textsc{TED} shows that prompted anomaly localization is not only about making models more sensitive to defects, but also about checking whether a high-scoring patch is a true defect or a hard normal region.
Our contributions are:
\begin{itemize}

  \item We identify a common transfer failure in CLIP-based anomaly localization: adapted models can give high anomaly scores to both true defects and visually complex normal regions, making them hard to rank apart.
  \item We show that the normal-versus-anomaly text response is still useful, but not precise enough: it often separates defects from ordinary background, while still confusing defects with hard normal regions.
  \item We propose \textsc{TED}, a post-hoc scoring method that compares ambiguous patches with source defect and hard-FP examples, with train-free and source-calibrated variants.

\end{itemize}

\section{Related Work}
\label{sec:related}

\paragraph{Backbones, dense representations, and anomaly readouts.}
Industrial anomaly detection has long relied on pretrained visual backbones, with surveys covering feature-, reconstruction-, density-, and distillation-based pipelines~\cite{liu2024deep,cui2023survey,lee2026continual}.
Feature matching methods such as SPADE~\cite{cohen2005sub_feature}, PaDiM~\cite{defard2021padim_feature}, PatchCore~\cite{roth2022patchcore_feature}, and pretrained-feature distribution modeling~\cite{rippel2021modeling_feautre} compare test patches with normal feature statistics or nearest-neighbor structures.
Other methods score anomalies through patch-distance criteria~\cite{ma2025patch}, normalizing flows such as FastFlow~\cite{yu2021fastflow}, CFlow-AD~\cite{gudovskiy2022cflow}, CS-Flow~\cite{rudolph2022fully_flow}, and SANFlow~\cite{kim2023sanflow}, reconstruction discrepancies~\cite{zavrtanik2021draem,hou2021divide_recon,ristea2022self_recon}, or teacher-student feature differences~\cite{bergmann2020uninformed_distill,deng2022anomaly_distill,wang2021student}.
For transformer backbones, layer selection, feature aggregation, and attention-sink mitigation are often needed for stable dense predictions~\cite{heckler2023exploring,zhang2024realnet,zhou2021deepvit_weak}.
These works indicate that localization quality depends not only on the representation, but also on how local evidence is scored and ranked.
\textsc{TED} follows this readout-centered view in prompted VLM pipelines, where visually complex normal regions can receive anomaly-like local responses.

\paragraph{Prompted vision-language anomaly detection.}
Vision-language models, especially CLIP~\cite{radford2021learning_clip}, provide a semantic image-text space for prompt-based anomaly detection and have motivated broader studies of VLM transfer~\cite{zhang2024vision,zhou2022learning_vlm}.
WinCLIP~\cite{jeong2023winclip} uses normal and anomalous prompt ensembles with window-level harmonic scoring for zero-shot pixel-level localization, while later work explores generalized prompts and anomaly-aware CLIP variants for data-efficient inspection~\cite{kim2025genclip,ma2025aa}.
Recent methods improve defect sensitivity through prompt learning or lightweight adaptation: AnomalyCLIP~\cite{zhou2023anomalyclip} learns object-agnostic prompts, AdaCLIP~\cite{cao2024adaclip} combines static and image-conditioned dynamic prompts, FAPrompt~\cite{zhu2025fine_faprompt} introduces fine-grained abnormality prompts, and AdaptCLIP~\cite{gao2025adaptclip} combines textual adaptation with visual and prompt-query adapters.
These methods improve benchmark performance, but stronger defect sensitivity does not necessarily yield cleaner local evidence: hard normal regions can still be ranked close to true defects.
Rather than adding another prompt learner or adapter, \textsc{TED} keeps the host response and changes how ambiguous local evidence is decoded using source defect and hard-FP support.

\paragraph{Backbone flexibility and multimodal anomaly understanding.}
Many CLIP-AD adaptation pipelines depend on CLIP-family tokenization, text encoders, prompt learners, feature hooks, layer choices, or host-specific scoring recipes.
Although they are not tied to a single image encoder, feature-layer changes or transfer to heterogeneous VLM backbones can still require re-engineering, retraining, or architecture search.
\textsc{TED} only needs patch-level visual features and a normal-versus-anomaly text response.
This lets the same source-evidence comparison work as a train-free score for raw VLM backbones such as ImageBind~\cite{girdhar2023imagebind}, or as a source-calibrated residual for adapted CLIP-AD hosts.
This differs from multimodal large-language-model approaches such as AnomalyGPT~\cite{gu2024anomalygpt}, zero-shot anomaly reasoning with MLLMs~\cite{xu2025towards}, MMAD~\cite{jiang2024mmad}, OmniAD~\cite{zhao2025omniad}, AnomalyR1~\cite{chao2025anomalyr1}, and AD-FM~\cite{liao2025adfm}, which target reasoning, explanation, instruction following, or end-to-end multimodal decisions with heavier architectures.
\textsc{TED} instead serves as a lightweight local evidence-decoding layer for the prompted VLM anomaly-localization pipelines evaluated here.


\begin{figure}[t]
\centering
\begin{subfigure}[t]{0.49\linewidth}
\centering
\includegraphics[width=\linewidth]{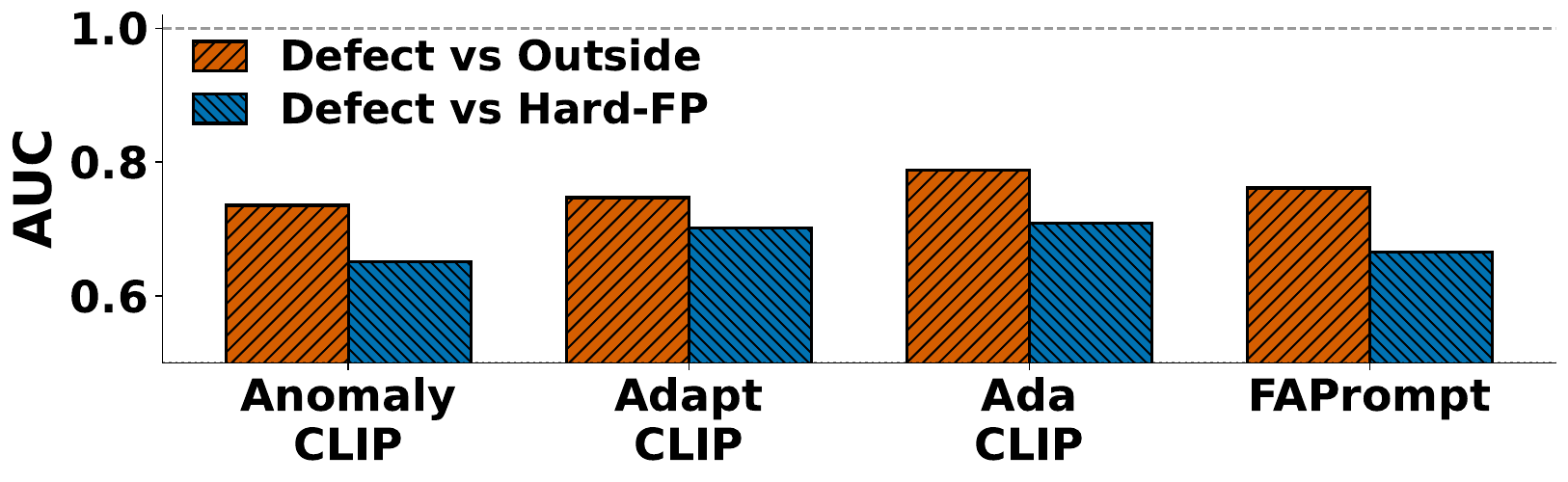}
\caption{Hard-FP competition.}
\end{subfigure}
\hfill
\begin{subfigure}[t]{0.49\linewidth}
\centering
\includegraphics[width=\linewidth]{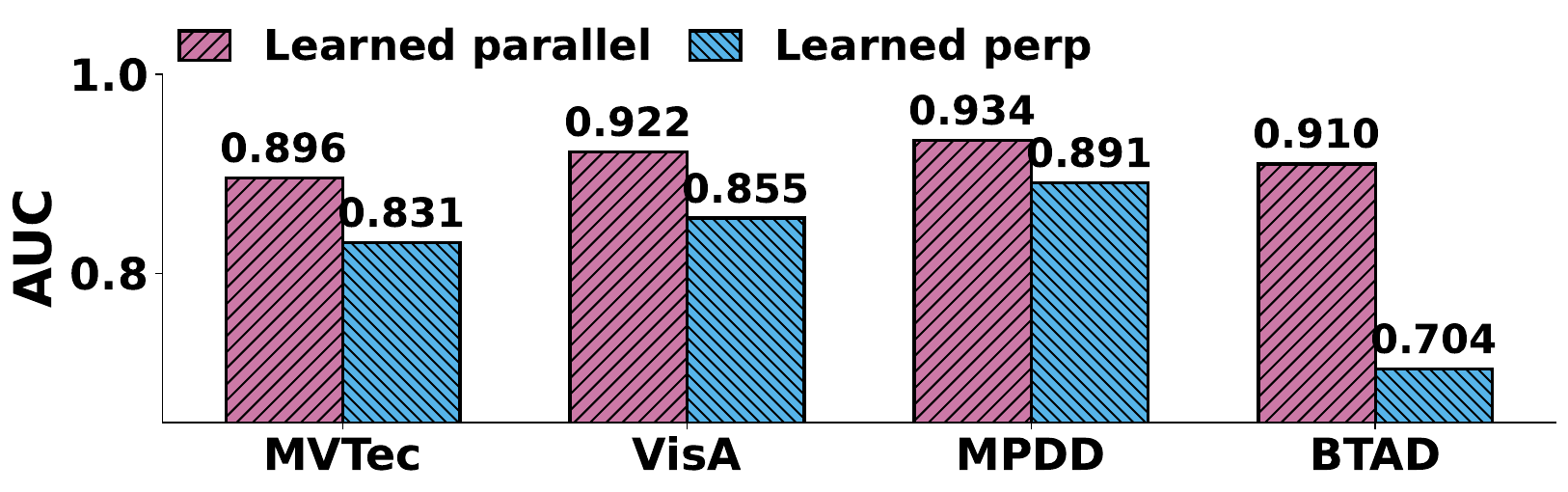}
\caption{Text-axis response.}
\end{subfigure}

\caption{
\textbf{Hard-FP competition and text-axis response.}
(a) Across adapted CLIP-AD hosts, true defects remain much harder to separate from hard false positives than from generic outside patches.
(b) Across target datasets, the learned text-axis response is consistently more informative than the orthogonal component, suggesting that the axis should be decomposed rather than discarded.
}
\label{fig:discuss1}
\end{figure}
\section{Proposed Method}
\label{sec:method}

\subsection{From Hard-FP Competition to Evidence Decomposition}
\label{sec:method_motivation}

We start from a simple failure case.
After adaptation, a CLIP-based anomaly detector may give high scores to true defects, but it can also give high scores to normal regions that only look suspicious.
We call these confusing normal regions \emph{hard false positives}.
They often come from strong edges, repeated textures, reflections, or salient object parts.
To measure this failure, we compare three kinds of patches: true defect patches, ordinary normal patches, and hard false-positive patches.
Fig.~\ref{fig:discuss1}(a) shows that adapted hosts separate true defects from ordinary normal patches more easily than from hard false positives.
Fig.~\ref{fig:discuss1}(b) shows that the host's normal-versus-anomaly text response is still useful, but not enough to solve this confusion.
In other words, the model has useful anomaly information, but its local score can still treat true defects and hard normal regions as similarly anomalous.

This leads to the main idea of \textsc{TED}.
A high anomaly score should not be trusted by itself.
Instead, we ask whether that high response looks more like source defect examples or source hard false-positive examples.
\textsc{TED} keeps the host's normal-versus-anomaly response, but re-ranks ambiguous patches using this source comparison.

\subsection{\textsc{TED} Overview}
\label{sec:method_ted}

\begin{figure*}[t]
\centering
\captionsetup[subfigure]{font=small,skip=2pt,justification=centering}

\begin{subfigure}[t]{0.32\textwidth}
  \centering
  \includegraphics[width=\linewidth]{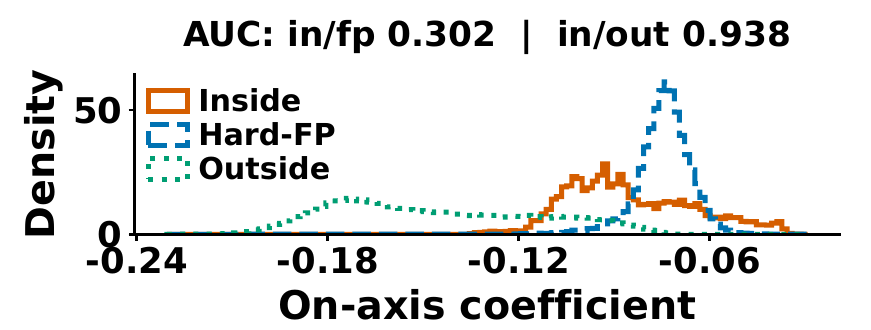}
  \caption{On-axis response.}
\end{subfigure}
\hfill
\begin{subfigure}[t]{0.32\textwidth}
  \centering
  \includegraphics[width=\linewidth]{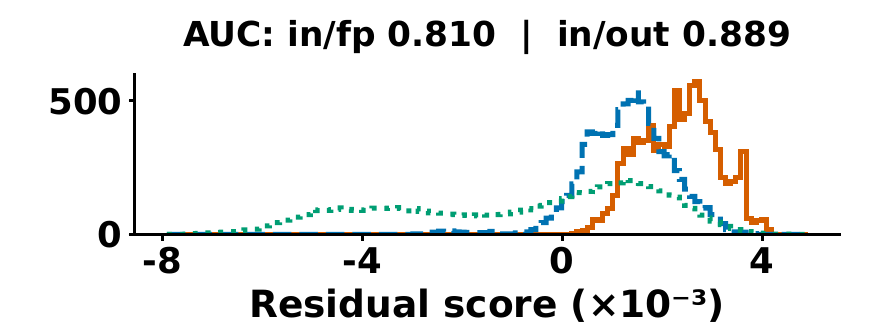}
  \caption{Orthogonal residual.}
\end{subfigure}
\hfill
\begin{subfigure}[t]{0.32\textwidth}
  \centering
  \includegraphics[width=\linewidth]{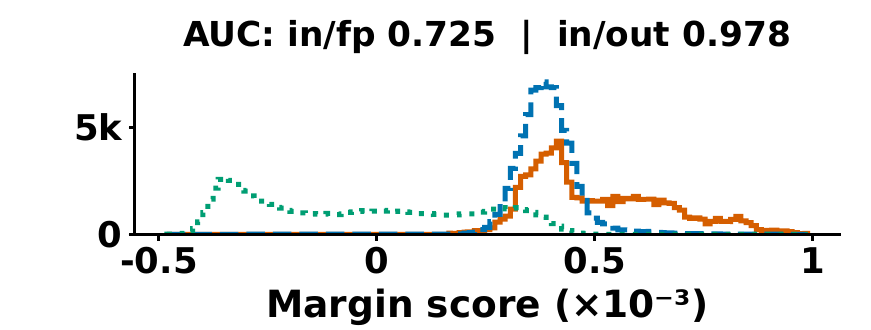}
  \caption{Source-supported margin.}
\end{subfigure}

\vspace{-0.4em}
\caption{
\textbf{Text-response entanglement and source-supported re-ranking.}
Raw on-axis scoring strongly overlaps true defects with hard false positives.
The orthogonal residual improves defect-vs-hard-FP separation in this case, but weakens separation from generic outside regions.
The source-supported margin re-ranks the same responses while preserving strong outside separation, showing that TED decomposes rather than discards the text-guided response.
}
\label{fig:discuss2}
\end{figure*}

\textsc{TED} is a post-hoc scoring method for anomaly localization.
It does not change the backbone, prompts, or host detector.
For each suspicious patch, \textsc{TED} asks whether it looks more like source defects or source hard false positives.
Defect-like patches are boosted, while hard-FP-like patches are suppressed.
T-TED uses this as the local score; C-TED uses it as a small correction to the adapted host score.
Neither mode uses target images, masks, labels, or target-score selection.
Fig.~\ref{fig:discuss2} shows this effect: raw scoring mixes defects with hard false positives, while \textsc{TED} re-ranks them using source examples.

\begin{figure}[t]
    \centering
    \includegraphics[width=0.9\linewidth]{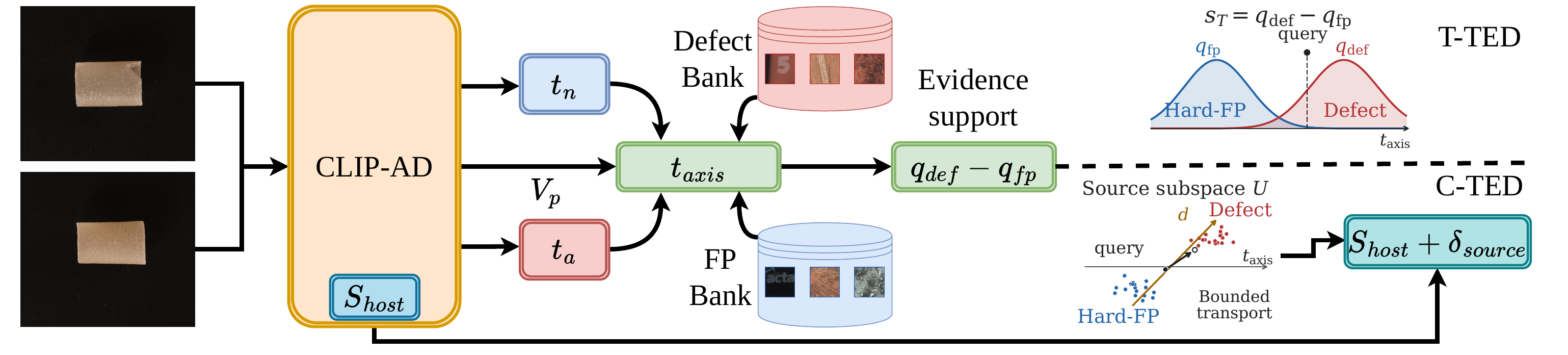}
    \caption{
    \textbf{Overview of \textsc{TED}.}
    The host provides patch features($V_{p}$) and normal$(t_n)$/anomaly$(t_a)$ text embeddings that define the host text axis.
\textsc{TED} projects query patches and source defect/hard-FP features onto this axis to compute a defect-versus-hard-normal support margin.
T-TED uses this margin as a train-free local score, while C-TED converts it into a bounded residual anchored to the host score.
    }
    \vspace{-1em}
    \label{fig:method}
\end{figure}

\subsection{Text-Axis Evidence Banks and Train-Free Score}
\label{sec:method_axis_bank}

For a query image $x$, the host provides normal/anomaly text embeddings $e_{\mathrm{n}}(x), e_{\mathrm{a}}(x)$ and patch features $v_i^{(\ell)}(x)$ at layer $\ell$.
We use the normal-to-anomaly text direction as a common one-dimensional scale:
\begin{equation}
u(x)=
\frac{e_{\mathrm{a}}(x)-e_{\mathrm{n}}(x)}
{\|e_{\mathrm{a}}(x)-e_{\mathrm{n}}(x)\|_2},
\qquad
c_i^{(\ell)}(x)=\langle v_i^{(\ell)}(x),u(x)\rangle .
\label{eq:text_axis_response}
\end{equation}
Here, $c_i^{(\ell)}(x)$ is the anomaly response of query patch $i$ on the host text direction.
\textsc{TED} does not modify this direction.
For each layer, we build two source banks:
$\mathcal{B}_{\mathrm{def}}^{(\ell)}=\{b_{\mathrm{def},j}^{(\ell)}\}_{j=1}^{N_{\mathrm{def}}}$ from source defect regions, and
$\mathcal{B}_{\mathrm{fp}}^{(\ell)}=\{b_{\mathrm{fp},j}^{(\ell)}\}_{j=1}^{N_{\mathrm{fp}}}$ from source normal patches that the host scores highly.
For either bank $z\in\{\mathrm{def},\mathrm{fp}\}$, each source patch is measured on the same text direction by $c_{z,j}^{(\ell)}(x)=\langle b_{z,j}^{(\ell)},u(x)\rangle$.
Thus, query patches, source defects, and source hard-FPs are compared on the same scale.
For each bank $z$, we compute how close the query response is to the bank responses:
\begin{equation}
q_z^{(\ell)}(v_i;x)
=
\log
\frac{1}{N_z}
\sum_{j=1}^{N_z}
\exp\left(
-\frac{(c_i^{(\ell)}(x)-c_{z,j}^{(\ell)}(x))^2}{\tau}
\right),
\qquad z\in\{\mathrm{def},\mathrm{fp}\}.
\label{eq:support_score}
\end{equation}
Here, $q_{\mathrm{def}}$ is large when the query looks like source defects, $q_{\mathrm{fp}}$ is large when it looks like source hard-FPs, and $\tau$ controls the similarity bandwidth.
The train-free \textsc{TED} score is
\begin{equation}
r_{\mathrm{TED}}^{(\ell)}(v_i;x)
=
q_{\mathrm{def}}^{(\ell)}(v_i;x)
-
\lambda_{\mathrm{fp}}
q_{\mathrm{fp}}^{(\ell)}(v_i;x).
\label{eq:ted_margin}
\end{equation}
A positive score means the patch is more defect-like; a negative score means it is more hard-FP-like.
We average this score over host-selected layers and use the resulting T-TED map in place of raw prompt similarity.

\subsection{Source-Calibrated \textsc{TED} Residual}
\label{sec:method_source_calibrated}

Adapted CLIP-AD hosts already have useful learned scores from prompts, adapters, fusion modules, or host-specific calibration.
Therefore, C-TED does not replace the host score.
Instead, it uses the source defect-versus-hard-FP margin to add a small bounded correction.
For each patch, we first compute the source margin
\begin{equation}
m_i^{(\ell)}
=
q_{\mathrm{def}}^{(\ell)}(v_i)
-
\lambda_{\mathrm{fp}}^{(\ell)}
q_{\mathrm{fp}}^{(\ell)}(v_i).
\label{eq:cted_margin}
\end{equation}
A large positive margin means the patch is closer to source defects; a negative margin means it is closer to source hard-FPs.
C-TED converts this margin into a bounded residual:
\begin{equation}
\Delta s_i^{(\ell)}
=
\gamma_\ell
\tanh(a_\ell \widehat{m}_i^{(\ell)} + b_\ell),
\label{eq:cted_residual}
\end{equation}
where $\widehat{m}_i^{(\ell)}$ is the normalized margin, and $a_\ell$, $b_\ell$, and $\gamma_\ell$ are learned only from source defect--hard-FP pairs.
The $\tanh$ term prevents the residual from becoming arbitrarily large.
The final C-TED score keeps the host score as the anchor:
\begin{equation}
s_{\mathrm{TED}}^{(\ell)}(v_i)
=
s_{\mathrm{host}}^{(\ell)}(v_i)
+
\Delta s_i^{(\ell)} .
\label{eq:cted_score}
\end{equation}
The residual is trained so that source defect patches score above source hard-FPs, while pairs already separated by the host are preserved.
After source training, all parameters are frozen.
No target images, masks, anomaly labels, scores, or metrics are used.



\begin{figure}[t]
\centering
\includegraphics[width=0.92\linewidth]{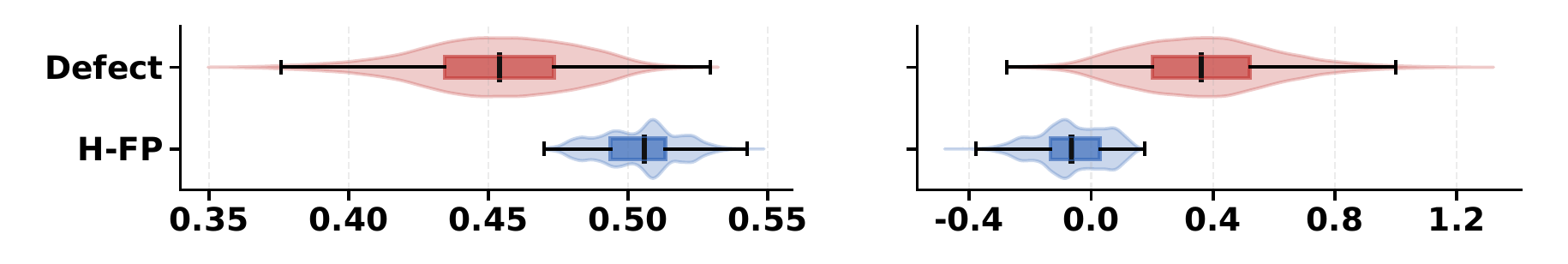}

\caption{
\textbf{Patch-level failure of raw prompt scoring.}
The left panel shows the raw prompt baseline and the right panel shows \textsc{TED}.
The baseline ranks hard-normal patches close to or above true defects, whereas \textsc{TED} re-ranks them using source defect and hard-FP support without backbone or prompt adaptation.
H-FP denotes hard false positive.
}
\label{fig:rawclip_hardfp_focus}
\end{figure}

\begin{figure}[t]
\centering
\includegraphics[width=0.95\linewidth]{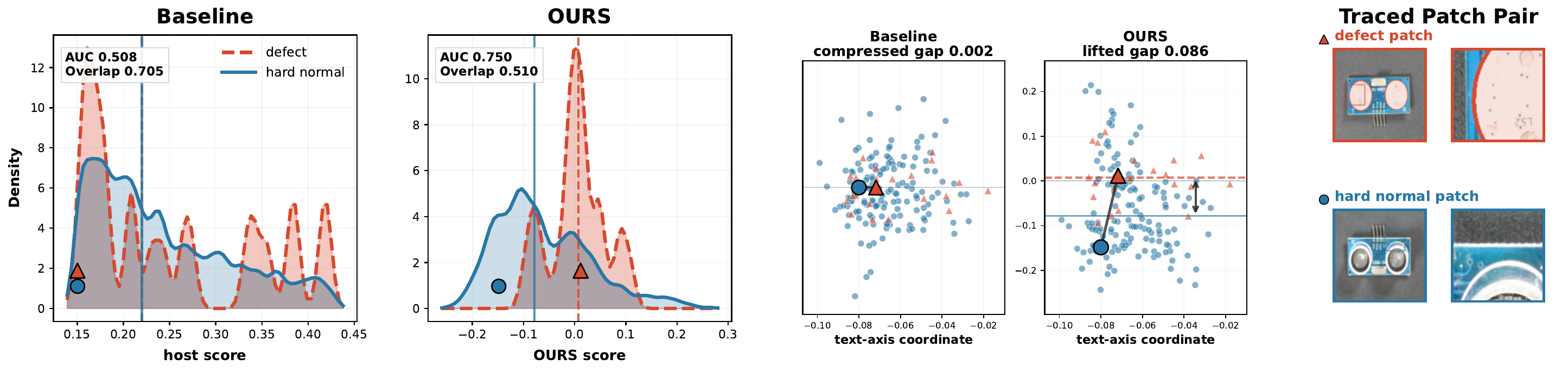}

\caption{
\textbf{Hard-FP decomposition in an adapted host.}
The baseline assigns overlapping high scores to true defect and hard-normal patches.
\textsc{TED} re-scores the same responses with source defect and hard-FP evidence, reducing overlap and lifting the defect--hard-normal ranking gap.
}
\label{fig:mainre}
\end{figure}

\begin{table*}[t]
\centering
\small
\setlength{\tabcolsep}{2.6pt}
\renewcommand{\arraystretch}{1.08}
\caption{
Cross-dataset transfer results for frozen VLM backbones with class-aware static prompts.
B denotes standard prompt similarity, T-TED train-free TED, and C-TED source-calibrated TED.
All variants use the same frozen backbone and prompts; only the local anomaly score changes.
Best results within each backbone and target dataset are bolded.
}
\label{tab:all_vlm}
\resizebox{\textwidth}{!}{%
\begin{tabular}{ll|cccc|cccc|cccc|cccc|cccc}
\toprule
\multirow{2}{*}{Backbone} & \multirow{2}{*}{Variant}
& \multicolumn{4}{c|}{MVTec\cite{bergmann2019mvtec}}
& \multicolumn{4}{c|}{VisA\cite{zou2022spot_visa}}
& \multicolumn{4}{c|}{MPDD\cite{liu2024deep}}
& \multicolumn{4}{c|}{BTAD\cite{mishra2021vt_btad}}
& \multicolumn{4}{c}{MVTec2\cite{heckler2026mvtec2}} \\
\cmidrule(lr){3-6} \cmidrule(lr){7-10} \cmidrule(lr){11-14} \cmidrule(lr){15-18} \cmidrule(lr){19-22}
& & I-AUC & P-AUC & PRO & AP
  & I-AUC & P-AUC & PRO & AP
  & I-AUC & P-AUC & PRO & AP
  & I-AUC & P-AUC & PRO & AP
  & I-AUC & P-AUC & PRO & AP \\
\midrule
\texttt{ViT-L/14-336} & B & \textbf{88.4} & 32.2 & 9.7 & 2.8 & 77.3 & 41.6 & 13.5 & 1.3 & 73.3 & 24.4 & 3.9 & 2.2 & 65.8 & 23.9 & 3.7 & 1.8 & 55.7 & 33.1 & 13.0 & 0.6 \\
 & T-TED & 88.2 & \textbf{69.3} & \textbf{64.0} & \textbf{21.2} & \textbf{79.4} & 59.2 & 42.3 & \textbf{6.7} & 78.5 & \textbf{76.4} & 60.5 & \textbf{5.4} & 71.3 & \textbf{77.9} & \textbf{50.8} & \textbf{17.2} & 58.4 & 68.2 & 31.5 & 6.4 \\
 & C-TED & 88.2 & 69.2 & 63.9 & 21.1 & \textbf{79.4} & \textbf{59.3} & \textbf{42.7} & 6.7 & \textbf{78.5} & 76.0 & \textbf{60.6} & 5.4 & \textbf{71.3} & 77.6 & 50.5 & 16.9 & \textbf{58.5} & \textbf{68.2} & \textbf{31.5} & \textbf{6.4} \\
\midrule
\texttt{ViT-B/16+} & B & 89.5 & 17.3 & 1.7 & 2.4 & 73.2 & 21.7 & 3.4 & 0.4 & \textbf{60.1} & 28.0 & 4.1 & 1.4 & 67.1 & 38.3 & 7.5 & 2.2 & \textbf{54.7} & 35.5 & 5.7 & 0.5 \\
 & T-TED & \textbf{90.0} & 82.9 & 67.8 & 17.1 & \textbf{75.5} & 77.4 & 51.9 & 6.1 & 59.4 & 70.8 & 48.4 & \textbf{10.2} & \textbf{69.0} & \textbf{63.3} & 34.8 & \textbf{6.1} & 54.5 & 64.5 & \textbf{29.7} & \textbf{2.7} \\
 & C-TED & \textbf{90.0} & \textbf{82.9} & \textbf{68.0} & \textbf{17.2} & \textbf{75.5} & \textbf{77.6} & \textbf{52.3} & \textbf{6.2} & 59.4 & \textbf{70.9} & \textbf{48.9} & 10.1 & \textbf{69.0} & 63.1 & \textbf{34.8} & 6.1 & 54.5 & \textbf{64.5} & 29.3 & 2.6 \\
\midrule
\texttt{ViT-L/14} & B & \textbf{85.3} & 30.7 & 8.2 & 2.7 & 68.1 & 44.6 & 16.3 & 1.3 & 71.6 & 25.1 & 2.8 & 2.0 & 61.0 & 22.8 & 2.7 & 1.8 & \textbf{59.1} & 36.8 & 12.9 & 0.6 \\
 & T-TED & 84.7 & \textbf{70.9} & \textbf{62.6} & \textbf{19.8} & \textbf{70.9} & 56.2 & 36.0 & \textbf{5.6} & \textbf{76.4} & \textbf{75.8} & \textbf{55.9} & \textbf{5.0} & \textbf{68.3} & \textbf{79.0} & \textbf{48.3} & \textbf{14.7} & 57.6 & 64.3 & 26.7 & 4.3 \\
 & C-TED & 84.7 & 70.9 & 62.4 & 19.7 & \textbf{70.9} & \textbf{56.3} & \textbf{36.1} & 5.6 & 76.4 & 75.6 & 55.9 & 5.0 & \textbf{68.3} & 78.5 & 47.7 & 14.2 & 57.6 & \textbf{64.3} & \textbf{26.9} & \textbf{4.3} \\
\midrule
\texttt{ViT-H/14} & B & 88.8 & 15.2 & 1.0 & 2.0 & 79.1 & 18.0 & 1.5 & 0.4 & \textbf{67.1} & 14.4 & 0.6 & 1.2 & 77.2 & 23.2 & 3.6 & 1.9 & 57.2 & 29.2 & 6.4 & 0.5 \\
 & T-TED & 90.4 & \textbf{85.3} & \textbf{77.9} & \textbf{26.2} & \textbf{79.4} & \textbf{82.3} & \textbf{61.8} & \textbf{9.5} & 65.5 & \textbf{87.6} & \textbf{77.2} & 19.0 & \textbf{84.5} & \textbf{78.3} & \textbf{55.0} & \textbf{13.9} & \textbf{60.4} & \textbf{71.8} & 38.9 & \textbf{4.7} \\
 & C-TED & \textbf{90.4} & 85.2 & 77.7 & 26.2 & \textbf{79.4} & 82.2 & 61.6 & 9.5 & 65.5 & 87.4 & 77.1 & \textbf{19.1} & \textbf{84.5} & 78.1 & 54.5 & 13.9 & \textbf{60.4} & 71.6 & \textbf{38.9} & 4.6 \\
\midrule
\texttt{ImageBind} & B & \textbf{90.5} & 14.4 & 0.7 & 2.0 & \textbf{79.7} & 16.5 & 1.0 & 0.4 & 65.2 & 19.5 & 1.7 & 1.3 & \textbf{86.1} & 18.6 & 2.9 & 1.8 & 58.0 & 26.9 & 3.9 & 0.4 \\
 & T-TED & 90.3 & \textbf{85.7} & \textbf{77.7} & 25.8 & 79.5 & \textbf{84.1} & \textbf{62.4} & \textbf{9.8} & \textbf{65.4} & \textbf{82.3} & \textbf{73.3} & 17.8 & 84.9 & \textbf{82.3} & \textbf{60.7} & \textbf{18.7} & 60.7 & \textbf{74.1} & \textbf{37.4} & \textbf{5.3} \\
 & C-TED & 90.3 & 85.6 & 77.7 & \textbf{25.9} & 79.6 & 84.0 & 62.3 & 9.8 & 65.4 & 82.1 & 73.1 & \textbf{17.9} & 84.9 & 82.1 & 60.3 & 18.6 & \textbf{60.7} & 73.9 & 37.4 & 5.1 \\
\bottomrule
\end{tabular}%
}
\end{table*}

\section{Experiments}
\label{sec:experiments}

\subsection{Experimental Setting}
\label{sec:exp_setting}

\paragraph{Protocol and metrics.}
We evaluate \textsc{TED} under a source-to-target anomaly localization protocol on MVTec AD, VisA, MPDD, and BTAD, with MVTec AD 2 used only for the frozen-backbone diagnostic.
Source data build evidence banks or train the source-calibrated residual; no target masks, anomaly labels, target scores, or target metrics are used for calibration or inference.
For class-aware baselines, class names only instantiate the fixed prompt ensemble and are not used for source-bank construction, calibration, residual-rank selection, insertion strength, model selection, or target-score selection.
Target masks are used only for evaluation.
We report I-AUROC and pixel-level P-AUROC, P-AP, and P-PRO, treating pixel-level localization as primary because \textsc{TED} modifies local anomaly maps.
I-AUROC is secondary, with official host global branches preserved when available.

\paragraph{Evaluation regimes.}
We evaluate three settings.
First, raw CLIP and ImageBind with static class-aware prompts test whether frozen multimodal features contain recoverable local defect evidence without anomaly-specific adaptation.
Second, adapted-host experiments keep each official CLIP-AD host unchanged and apply C-TED only as a post-hoc local evidence decoder.
Third, layer/backbone recipe diagnostics test whether hard-FP competition is simply a representation-choice artifact, using both fixed-host swaps and host retraining with rebuilt source banks.
\paragraph{Implementation details.}
For each host, we preserve its official prompts, text encoder, selected layers, image resolution, post-processing, and metrics whenever possible.
Defect banks are built from source anomalous patches overlapping source masks; hard-FP banks are built from source normal patches with high official host anomaly scores.
T-TED uses the signed support margin directly, whereas C-TED anchors the host local score and learns only a bounded source-domain residual from source defect and hard-FP pairs.
Unless otherwise stated, bank sizes, residual ranks, support bandwidths, and hard-FP mining rules are fixed before target evaluation; all variants use the same source-to-target split, and multi-seed results report mean and standard deviation.

\subsection{Main Results and Diagnostics}
\label{sec:exp_results}

\paragraph{Frozen-backbone evidence recovery.}
We first use raw VLM as a diagnostic setting for local evidence decoding.
Table~\ref{tab:all_vlm} evaluates frozen VLM backbones with class-aware static prompts, where B uses normal-versus-abnormal prompt similarity and T-TED/C-TED change only the local anomaly score.
Across backbones and targets, direct prompt similarity gives weak pixel-level localization, especially in P-PRO and P-AP, while \textsc{TED} substantially improves pixel metrics.
I-AUC changes are smaller and sometimes mixed, consistent with TED targeting local maps rather than image-level screening.
This contrast is important: the same frozen representation can support much better localization once the local response is decoded against source defect and hard-FP evidence.
In other words, the frozen backbone is not necessarily missing all defect information; the problem is that raw prompt similarity ranks that information poorly.
This also shows why \textsc{TED} does not need to update the backbone in this setting.
It changes how local responses are scored, not what visual features are extracted.
Fig.~\ref{fig:rawclip_hardfp_focus} shows the patch-level failure: raw prompt scoring can rank hard-normal patches above true defects, whereas \textsc{TED} reverses this ranking using source defect and hard-FP support without changing the backbone or prompts.
Together, Table~\ref{tab:all_vlm} and Fig.~\ref{fig:rawclip_hardfp_focus} show that frozen multimodal backbones can contain recoverable defect evidence, but direct prompt scoring may decode it poorly against hard-normal responses.
Thus, this diagnostic separates representation capacity from local readout quality, supporting the evidence-decoding view behind \textsc{TED}.

\begin{figure*}[t]
\centering
\captionsetup[subfigure]{skip=2pt}


  \begin{subfigure}[t]{0.24\textwidth}
  \centering
  \vspace{0pt}
  \includegraphics[height=2.8cm]{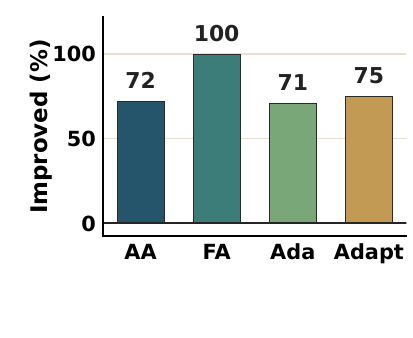}
  \vspace{-2.5em}
  \caption{Host-wise.}
  \end{subfigure}
  \hfill
  \begin{subfigure}[t]{0.24\textwidth}
  \centering
  \vspace{0pt}
  \includegraphics[height=2.8cm]{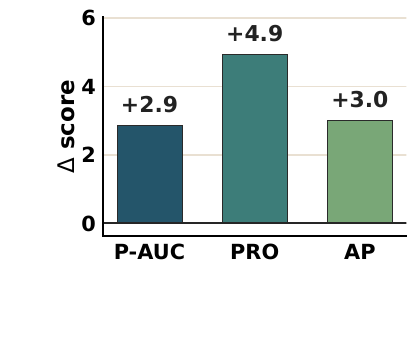}
  \vspace{-2.5em}
  \caption{Metric-wise.}
  \end{subfigure}
  \hfill
  \begin{subfigure}[t]{0.24\textwidth}
  \centering
  \vspace{0pt}
  \includegraphics[height=2.8cm]{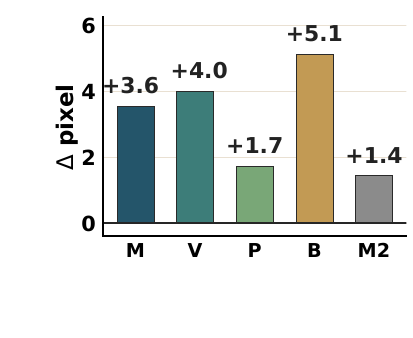}
  \vspace{-2.5em}
  \caption{Target-wise.}
  \end{subfigure}
  \hfill
  \begin{subfigure}[t]{0.24 \textwidth}
  \centering
  \vspace{0pt}
  \includegraphics[height=2.8cm]{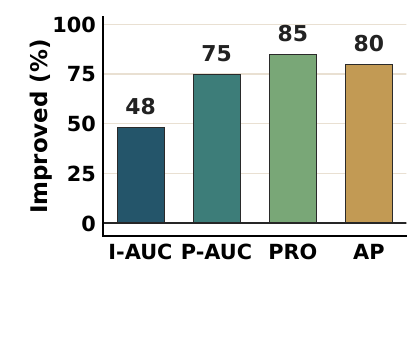}
  \vspace{-2.5em}
  \caption{Consistency.}
  \end{subfigure}

  \vspace{0.55em}

\begin{subfigure}[t]{0.28\textwidth}
\centering
\vspace{0pt}
\includegraphics[height=2.55cm]{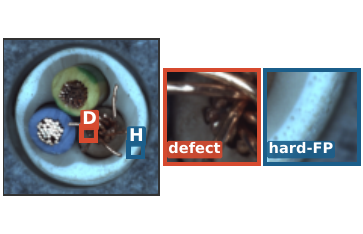}
\caption{Traced patches.}
\end{subfigure}
\hfill
\begin{subfigure}[t]{0.42\textwidth}
\centering
\vspace{0pt}
\includegraphics[height=2.55cm]{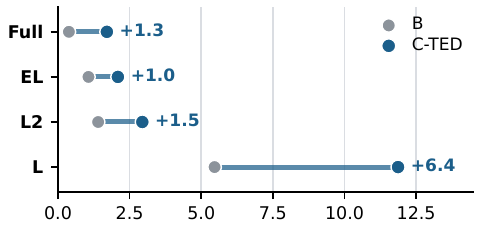}
\caption{Patch-rank margin.}
\end{subfigure}
\hfill
\begin{subfigure}[t]{0.26\textwidth}
\centering
\vspace{0pt}
\includegraphics[height=2.55cm]{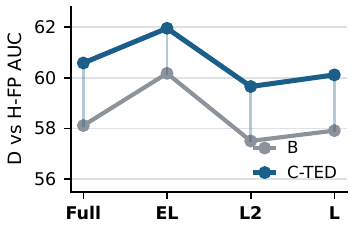}
\caption{Aggregate separation.}
\end{subfigure}

\caption{
\textbf{Adapted-host and recipe-retuning diagnostics.}
\textbf{Top}: source-calibrated TED improves local pixel ranking across adapted CLIP-AD hosts; M, V, P, B, and M2 denote MVTec AD , VisA, MPDD, BTAD, and MVTec AD 2, respectively.
\textbf{Bottom}: changing and retraining layer recipes shifts the host response, but the same traced defect and hard-FP patches remain competitive under the baseline.
C-TED increases the patch-rank margin and yields cleaner aggregate separation between defect-supported and hard-normal-supported responses.
}
\label{fig:adapted_host_recipe_combined}
\end{figure*}

\paragraph{Hard-FP decomposition in adapted hosts.}
Fig.~\ref{fig:mainre} illustrates an adapted-host case where true defect and hard-normal patches receive overlapping high anomaly scores.
\textsc{TED} re-scores the same ambiguous responses with source defect and hard-FP evidence, reducing their overlap and increasing the local ranking gap.
Fig.~\ref{fig:adapted_host_recipe_combined} (top) summarizes this behavior across adapted CLIP-AD hosts, target datasets, and pixel-level metrics.
C-TED improves most pixel-level settings, with gains distributed across P-AUC, P-PRO, and P-AP.
The gains are host-dependent: highly calibrated hosts leave less room for correction, while settings with stronger hard-FP competition benefit more.
This is expected because C-TED is designed to correct remaining hard-normal competition, not to overwrite the host's learned anomaly response.
When the baseline already separates true defects from hard-normal regions, the correction tends to be smaller.
When the two groups overlap, the source defect and hard-FP banks provide a clearer local re-ranking signal.
This pattern indicates that C-TED complements host adaptation by correcting residual local ranking errors rather than replacing the adapted detector.
All adapted-host corrections are estimated only from source-domain evidence banks; full host-wise and dataset-wise results are reported in the Appendix.

\begin{wraptable}{r}{0.4\columnwidth}
\centering
\caption{
\textbf{Failure-conditioned gains.}
Bins are grouped by baseline hard-FP severity before inspecting C-TED gains.
}
\label{tab:failure_conditioned_gains}
\scriptsize
\setlength{\tabcolsep}{2.0pt}
\renewcommand{\arraystretch}{1.03}
\resizebox{\linewidth}{!}{%
\begin{tabular}{lcccc}
\toprule
H-FP bin & \#bins & Sev. & $\Delta$PRO / $\Delta$AP & $\Delta$Loc \\
\midrule
Low  & 3 & 0.07 & +3.7 / +6.3  & +5.0 \\
Mid  & 3 & 0.10 & +8.8 / +13.8 & +11.3 \\
High & 3 & 0.19 & +11.3 / +9.8 & +10.6 \\
\bottomrule
\end{tabular}%
}
\end{wraptable}

\paragraph{Failure-conditioned gains.}
We test whether hard-FP severity predicts when \textsc{TED} helps.
If C-TED corrects hard-FP entanglement, gains should grow when the baseline assigns stronger anomaly evidence to hard-normal regions.
Table~\ref{tab:failure_conditioned_gains} groups completed FAPrompt/AdaCLIP target classes into Low, Mid, and High regimes using only baseline hard-normal responses.
This analysis-only grouping uses only baseline hard-normal responses, without C-TED scores, target metrics, calibration, tuning, or model selection.
C-TED improves all regimes, but mean localization gain over $\Delta$P-PRO and $\Delta$P-AP rises from $+5.0$ to $+10.9$ from Low to Mid/High.
This makes the result useful as a diagnostic: it checks whether the measured failure mode aligns with where C-TED improves most.
It also explains why we avoid presenting C-TED as a generic post-processing boost.
The grouping is not used to choose any parameter, so the table should be read as post-hoc evidence for the mechanism, not as a tuning rule.
Thus, \textsc{TED} is not a uniform score shift; it helps most when anomaly-sensitive hosts rank hard-normal evidence close to true defects.
This trend is consistent with smaller gains in already well-separated settings, where less hard-FP competition remains to correct.

\begin{figure*}[t]
  \centering
  \captionsetup[subfigure]{skip=2pt}

  \begin{subfigure}[t]{0.38\textwidth}
  \centering
  \includegraphics[height=2.95cm]{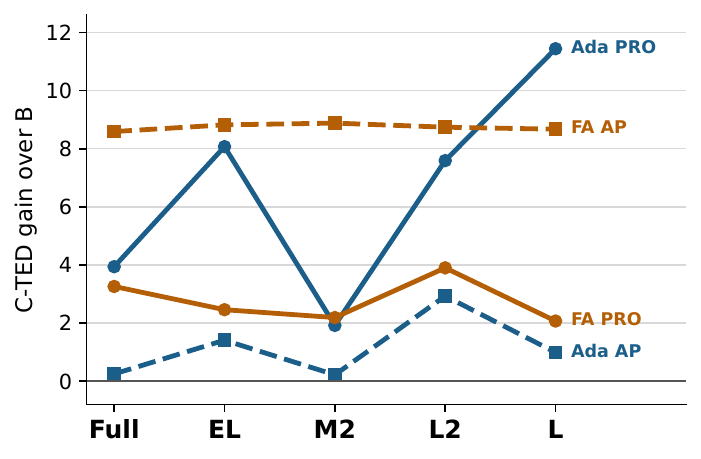}
  \caption{Layer retuning.}
  \end{subfigure}
  \hfill
  \begin{subfigure}[t]{0.30\textwidth}
  \centering
  \includegraphics[height=2.95cm]{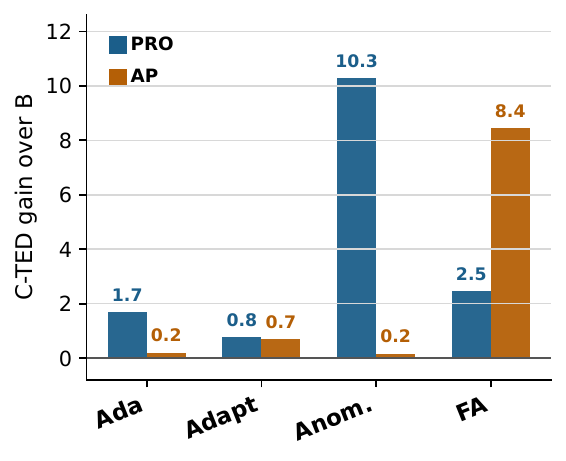}
  \caption{Backbone retuning.}
  \end{subfigure}
  \hfill
  \raisebox{2.5cm}{%
  \begin{minipage}[t]{0.27\textwidth}
  \centering
  \footnotesize
  \textbf{Retuning setting}
  \vspace{2pt}

  \scriptsize
  \setlength{\tabcolsep}{1.8pt}
  \renewcommand{\arraystretch}{1.05}
  \resizebox{\linewidth}{!}{%
  \begin{tabular}{lccc}
  \toprule
  Setting & Retrain & Bank & PRO / AP \\
  \midrule
  Fixed swap & -- & \textbf{V} & $+\!3.6$ / $+\!2.8$ \\
  Layer recipe & \textbf{V} & \textbf{V} & $+\!4.7$ / $+\!5.0$ \\
  Backbone recipe & \textbf{V} & \textbf{V} & $+\!3.8$ / $+\!2.4$ \\
  \bottomrule
  \end{tabular}
  }

  \vspace{1pt}
  \scriptsize
  Gain over baseline; banks are source-only.
  \end{minipage}
  }

  \vspace{0.15em}

  \begin{subfigure}[t]{0.35\textwidth}
  \centering
  \includegraphics[height=2.90cm]{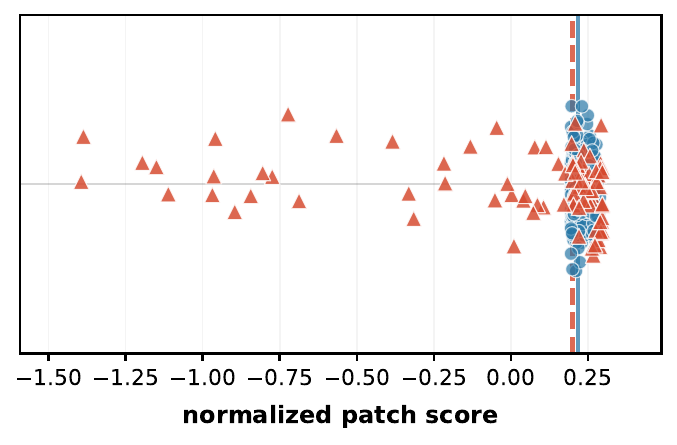}
  \caption{Defect-only support.}
  \end{subfigure}
  \hfill
  \begin{subfigure}[t]{0.35\textwidth}
  \centering
  \includegraphics[height=2.90cm]{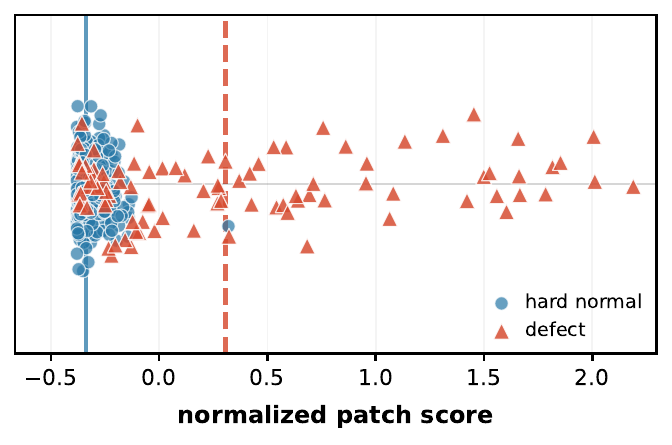}
  \caption{With hard-FP support.}
  \end{subfigure}
  \hfill
  \begin{subfigure}[t]{0.24\textwidth}
  \centering
  \includegraphics[height=2.90cm]{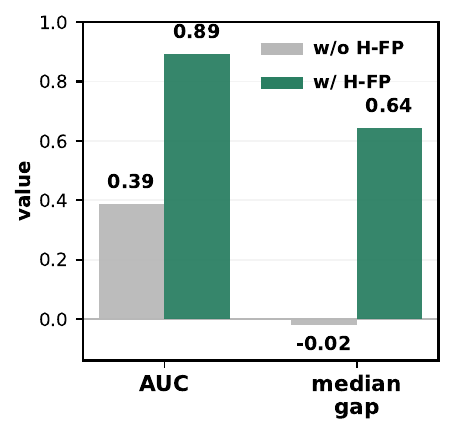}
  \caption{Separation gain.}
  \end{subfigure}

  \caption{
  \textbf{Recipe-consistent retuning and hard-FP support.}
  \textbf{Top}: C-TED remains beneficial after changing the feature-layer or backbone recipe, retraining the host when applicable, and rebuilding source banks for the resulting representation.
  \textbf{Bottom}: defect-only support leaves defect and hard-normal patches entangled, whereas adding hard-FP support explicitly models the competing hard-normal evidence and produces a cleaner local ranking
  margin.
  }
  \label{fig:recipe_hardfp_combined}
  \end{figure*}

\paragraph{Recipe-consistent retuning analysis.}
A natural concern is that hard-FP competition reflects a poor layer or backbone recipe.
We test recipe-consistent settings where the host is retrained after recipe changes and source banks are rebuilt.
Fig.~\ref{fig:adapted_host_recipe_combined} (bottom) traces the same defect and hard-FP pair across retrained layer recipes.
Although recipes shift host responses, the baseline still ranks the defect and hard-FP pair closely.
In other words, changing the layer recipe can move the scores, but it does not always make the defect clearly outrank the confusing normal patch.
This is why recipe search and evidence decoding address different parts of the problem.
Recipe search changes which features the host uses, while C-TED changes how ambiguous high-scoring patches are compared against source defect and hard-FP examples.
Fig.~\ref{fig:recipe_hardfp_combined} (top) further shows that C-TED improves local PRO/AP after layer/backbone retraining and source-bank reconstruction.
Together, these results suggest that recipe selection alone cannot ensure clean local ranking.

\subsection{Ablation and Control Studies}
\label{sec:exp_ablation}

\paragraph{Evidence components and residual rank.}
Table~\ref{tab:ted_ablation} ablates source evidence and residual rank.
On adapted hosts, T-TED replacement can be brittle because the official score is already calibrated; C-TED instead anchors the host map with a bounded source-calibrated residual.
Hard-FP support helps most under strong hard-normal competition, and the rank ablation shows that a small residual subspace is sufficient.
Fig.~\ref{fig:recipe_hardfp_combined} (bottom) confirms that hard-FP support separates defect and hard-normal patches more clearly than defect-only support.
This means the useful signal comes from modeling the confusing normal patches, not simply from adding more correction capacity.
Together, these results support hard-FP-aware residual correction over host-score replacement or larger residual capacity.

\begin{table*}[t]
  \centering
  \caption{
  Ablation study of \textsc{TED} on BTAD.
  (a) validates the calibrated decoding components on adapted hosts, and
  (b) analyzes the effect of the residual rank on AA-CLIP.
  }
  \vspace{-1em}
  \label{tab:ted_ablation}
  \scriptsize
  \setlength{\tabcolsep}{4pt}

  \begin{minipage}[t]{0.51\textwidth}
  \centering
  \vspace{3pt}
  \renewcommand{\arraystretch}{1.08}
  \resizebox{\linewidth}{!}{
  \begin{tabular}{@{}llccc@{}}
  \toprule
  Host & Method & P-AUROC & P-PRO & P-AP \\
  \midrule
  \multirow{4}{*}{AA-CLIP}
  & Baseline & 92.48 & 59.78 & 36.55 \\
  & T-TED & 86.58 & 53.50 & 27.88 \\
  & C-TED w/o hard-FP & 93.58 & 61.76 & 38.40 \\
  & C-TED & \textbf{94.23} & \textbf{65.42} & \textbf{39.68} \\
  \midrule
  \multirow{4}{*}{FAPrompt}
  & Baseline & 90.52 & 60.32 & 21.54 \\
  & T-TED & 90.00 & 60.12 & 20.33 \\
  & C-TED w/o hard-FP & 94.99 & 70.93 & 45.01 \\
  & C-TED & \textbf{94.99} & \textbf{70.95} & \textbf{45.01} \\
  \bottomrule
  \end{tabular}
  }
  \\
  \textbf{(a) Component ablation}
  \end{minipage}
  \hfill
  \begin{minipage}[t]{0.47\textwidth}
  \centering
  \vspace{3pt}

  \renewcommand{\arraystretch}{1.8}
  \resizebox{\linewidth}{!}{
  \begin{tabular}{@{}lcccc@{}}
  \toprule
  Rank & I-AUROC & P-AUROC & P-PRO & P-AP \\
  \midrule
  Baseline & 90.62 & 92.48 & 59.78 & 36.55 \\
  $r=2$ & 91.10 & \textbf{94.23} & \textbf{65.42} & \textbf{39.68} \\
  $r=4$ & 91.26 & 93.83 & 65.06 & 38.99 \\
  $r=8$ & \textbf{91.46} & 93.51 & 63.67 & 38.26 \\
  \bottomrule
  \end{tabular}
  }
  \\
  \textbf{(b) Residual rank ablation}
  \end{minipage}
\end{table*}

\begin{table}[t]
\centering
\caption{
\textbf{Source-control ablations.}
Controls corrupt source supervision or replace TED with scalar source readouts using the same evidence.
Values are gains over the host baseline.
}
\label{tab:source_memory_controls}
\footnotesize
\setlength{\tabcolsep}{3.8pt}

\begin{minipage}[t]{0.52\linewidth}
\centering
\textbf{(a) Source-supervision semantics}
\resizebox{0.99\linewidth}{!}{
\begin{tabular}{@{}lcccc@{}}
\toprule
Control & $n$ & $\Delta$P-AUC & $\Delta$PRO & $\Delta$AP \\
\midrule
Correct C-TED & 3 & \textbf{+0.12} & \textbf{+2.47} & \textbf{+2.07} \\
Label shuffle & 3 & -1.42 & -1.16 & -2.12 \\
Label swap & 3 & -6.29 & -7.93 & -9.32 \\
\bottomrule
\end{tabular}
}
\end{minipage}
\hfill
\begin{minipage}[t]{0.47\linewidth}
\centering
\textbf{(b) Scalar source readout}
\vspace{2pt}

\resizebox{0.95\linewidth}{!}{
\begin{tabular}{@{}lccc@{}}
\toprule
Control & $n$ & $\Delta$PRO / $\Delta$AP & $\Delta$Loc \\
\midrule
  Scalar replace  & 16 & -19.8 / -14.1 & -17.0 \\
  Scalar residual & 16 & +2.1 / -5.3  & -1.6 \\
  C-TED           & 16 & \textbf{+4.4} / \textbf{+3.7} & \textbf{+4.1} \\
\bottomrule
\end{tabular}
}
\end{minipage}
\end{table}

\paragraph{Source-supervision controls.}
Table~\ref{tab:source_memory_controls} provides source-only controls for the residual correction.
Label shuffle and label swap keep the same source split, bank size, residual capacity, and calibration budget, but corrupt the defect-versus-hard-FP assignment.
These variants reduce or remove the gain, suggesting that the correction depends on the semantic direction of the source evidence, not only on added residual
capacity.

The second control trains simple source-supervised scalar readouts on the same source defect and hard-FP banks, either replacing the local map or adding a
residual to the host map.
These readouts are weaker than C-TED in the completed settings.
The Appendix further tests host-score-only calibrators and additional learned readouts, showing that scalar calibration is not a clean substitute for source-
supported evidence comparison.
Thus, the gains are not fully explained by source-bank access, scalar supervision, or host-score recalibration alone.
TED uses source banks to contrast defect and hard-FP support
along the host's text axis, rather than treating full-dimensional
visual similarity as the anomaly criterion.
Additional same-bank controls compare full-dimensional 1-NN,
prototype, and LogMeanExp readouts in both replacement and
residual modes (Appendix~B.6,
Table~\ref{tab:same_bank_controls}).
Across the 18 evaluated settings, none of these Full-D readouts
improves all three pixel-level metrics on average, whereas
C-TED does.

\begin{wraptable}{r}{0.32\columnwidth}
\centering
\vspace{-1em}
\caption{\textbf{Fixed-bank sensitivity.}}
\label{tab:efficiency_bank_sensitivity}
{\fontsize{5.8pt}{6.2pt}\selectfont
\setlength{\tabcolsep}{1.6pt}
\renewcommand{\arraystretch}{0.86}
\resizebox{0.96\linewidth}{!}{%
\begin{tabular}{@{}ccccc@{}}
\toprule
Bank & Ret. & $\Delta$PRO & $\Delta$AP & $\Delta$Loc \\
\midrule
64  & 256  & +7.17 & +13.33 & +10.25 \\
128 & 512  & +7.12 & +13.32 & +10.22 \\
512 & 2048 & +7.18 & +13.33 & +10.25 \\
\bottomrule
\end{tabular}%
}
}
\end{wraptable}

\paragraph{Efficiency and fixed-policy sensitivity.}
Table~\ref{tab:efficiency_bank_sensitivity} varies the retained source-bank size while keeping the source split, target protocol, resolution, and post-processing fixed.
Across tested budgets, C-TED yields nearly identical localization gains, indicating that the correction saturates with a moderate retained bank size.
This reduces sensitivity to the exact retained-bank budget and avoids selecting bank sizes from target-domain performance.
Because source banks are built once from the source split and then reused, the evaluation does not introduce target-specific bank construction.
These results support using a fixed source-bank policy across target datasets without target-domain score selection, per-target bank retuning, or target-specific threshold search.
\section{Conclusion}
We presented \textsc{TED}, a text-axis evidence decomposition framework for prompted anomaly localization.
Our analysis showed that adapted CLIP-based anomaly detectors can remain anomaly-sensitive while ranking true defects close to visually complex normal regions.
Rather than replacing the host detector or discarding its text-guided response, \textsc{TED} re-ranks ambiguous local responses by comparing source defect support with source hard-false-positive support.
This yields a train-free score for raw VLM backbones and a source-calibrated residual correction for adapted CLIP-AD hosts, while leaving the visual backbone and prompts unchanged and requiring no target-domain training.
Across frozen multimodal backbones, adapted hosts, and cross-domain transfer settings, \textsc{TED} improves pixel-level localization by reducing hard-FP competition and recovering defect evidence obscured by raw prompt similarity or host-specific local scoring.
These results suggest that prompted anomaly localization should be treated not only as an adaptation problem, but also as a local evidence-decoding problem, with
gains depending on remaining hard-FP competition and representative source evidence.
\begin{ack}
This work was supported by the Institute of Information
\& Communications Technology Planning \& Evaluation (IITP)
grant funded by the Korea government (MSIT)
[RS-2021-II211341, Artificial Intelligence Graduate School
Program (Chung-Ang University), and RS-2022-II220124,
Development of Artificial Intelligence Technology for
Self-Improving Competency-Aware Learning Capabilities].
This research was also supported by the AI Seoul Tech
Research Support Program of the Seoul Future Foundation.
\end{ack}



\bibliographystyle{plain}
\bibliography{main}

\newpage
\appendix

\section{Additional Experimental Details}
\label{sec:supp_exp_details}

\subsection{Implementation Details}
\label{subsec:supp_implementation}

For each adapted CLIP-AD host, we keep the official visual backbone, text encoder, prompt format, selected feature layers, image resolution, post-processing, and metric implementation unchanged whenever possible.
\textsc{TED} is applied only as a post-hoc local evidence decoder.
For T-TED, the signed source-support margin is used directly as the local anomaly map.
For C-TED, the official host local score remains the anchor, and a bounded source-calibrated residual adjusts the local ranking.

The source defect bank is constructed from source-domain anomalous patches whose spatial support overlaps source anomaly masks.
The hard-FP bank is constructed from source normal patches that receive high local anomaly scores under the corresponding host baseline.
Both banks are stored as visual features and projected onto the query-specific normal-versus-anomaly text response at inference time.
For backbone-swap or recipe-retuning experiments, source banks are rebuilt for the resulting feature representation.
No target images, target masks, target anomaly labels, target scores, or target metrics are used to select bank entries, calibration parameters, residual ranks, insertion strengths, or support bandwidths.
Class names, when required by the fixed prompt protocol, are used only for prompt instantiation.

The C-TED residual calibrator is trained only from source defect and source hard-FP pairs.
The calibration objective encourages source defect patches to rank above source hard-FP patches after correction, while preserving source pairs already separated by the host.
All calibration parameters are frozen before target evaluation.
Unless otherwise stated, the same fixed bank policy and hyperparameters are used across target datasets.

\subsection{Fixed Evaluation Policy}
\label{subsec:supp_fixed_policy}

Table~\ref{tab:supp_fixed_policy} summarizes the fixed source-calibration policy.
All entries are determined before target evaluation.
No target-domain mask, anomaly label, target score, or target metric is used for policy selection.
Class names are used only for fixed prompt instantiation when required by the corresponding baseline protocol, and not for source-bank construction, calibration, residual-rank selection, insertion strength, support bandwidth, hard-FP mining, or model selection.

\begin{table}[t]
\centering
\caption{
\textbf{Fixed evaluation policy.}
All entries are fixed before target evaluation and are not selected using target-domain scores or masks.
}
\label{tab:supp_fixed_policy}
\scriptsize
\setlength{\tabcolsep}{3.4pt}
\renewcommand{\arraystretch}{1.06}
\begin{tabular}{ll}
\toprule
Item & Policy \\
\midrule
Defect bank & Source anomalous patches overlapping source masks \\
Hard-FP bank & High-scoring source normal patches under the host baseline \\
Target usage & Target masks and scores are used only for final evaluation \\
Bank budget & Fixed per host; not selected per target dataset \\
Hard-FP mining & Fixed source-normal top fraction \\
Support bandwidth & Fixed before target evaluation \\
FP weight $\lambda_{\mathrm{fp}}$ & Fixed before target evaluation \\
Residual rank $r$ & Fixed source-calibrated rank \\
Calibration data & Source defect and source hard-FP pairs only \\
Calibration budget & Fixed epochs, optimizer, and train-point budget \\
Host components & Backbone, prompts, layers, and post-processing unchanged \\
\bottomrule
\end{tabular}
\end{table}

\subsection{Dataset, Metrics, and Compute}
\label{subsec:supp_dataset_metric_compute}

We evaluate cross-domain anomaly localization on MVTec AD, VisA, MPDD, and BTAD.
MVTec AD contains industrial object and texture categories with pixel-level anomaly masks; VisA contains object-centric industrial inspection categories with diverse normal structures; MPDD contains metal-part defects with challenging texture and shape variations; BTAD contains three industrial categories with pixel-level annotations and strong class imbalance.
MVTec AD 2 is additionally used as a held-out diagnostic target for frozen-backbone and adapted-host stress evaluations when completed.

We report image-level AUROC (I-AUROC) and pixel-level AUROC (P-AUROC), average precision (P-AP), and AUPRO (P-PRO).
Since \textsc{TED} modifies local anomaly maps rather than the global image-level branch, pixel-level metrics are the primary target, and I-AUROC is reported for transparency.
For every host and backbone, the baseline and \textsc{TED} variants use the same image resolution, post-processing, and metric implementation.

All experiments were run on NVIDIA RTX 6000 Ada GPUs with 48GB memory.
Most jobs use a single GPU; parallel sweeps run independent host/transfer settings on separate GPUs without changing the evaluation protocol.
For retained-bank sensitivity, peak GPU memory is approximately 3.8--3.9GB for AA-CLIP and 4.1GB for FAPrompt.
Source banks are constructed once from the source split and reused at inference time, so the added cost scales with the retained bank budget.

\section{Additional Ablation and Control Studies}
\label{sec:supp_ablations}

\subsection{Training-Control Sanity Checks}
\label{subsec:supp_training_control}

Table~\ref{tab:cted_training_control} tests whether C-TED's gain can be explained by extra source-only residual capacity alone.
All variants use the same source-only calibration budget and residual capacity.
On AA-CLIP MVTec$\rightarrow$BTAD, correct C-TED improves P-AUC/PRO/AP by $+1.03/+3.88/+1.71$, whereas label swap gives $-6.40/-8.28/-10.45$.
On AA-CLIP VisA$\rightarrow$BTAD, C-TED is mixed in P-AUC ($-0.98$) but improves PRO/AP by $+0.69/+4.09$; label swap degrades all three metrics by $-12.19/-15.95/-17.23$.
On AdaCLIP VisA$\rightarrow$BTAD, correct C-TED gives smaller positive changes of $+0.31/+2.84/+0.41$.
The detailed AA-CLIP controls further show that label shuffle degrades AP by $-3.95$, and a random residual subspace gives a smaller AP gain ($+0.59$) than correct C-TED ($+1.71$).
These controls support the modest conclusion that the source defect-versus-hard-FP direction matters, rather than residual capacity alone.

\begin{table*}[t]
\centering
\caption{
\textbf{Training-control sanity checks for C-TED.}
All variants use the same source-only calibration budget and residual capacity.
Panel (a) tests the effect of swapping defect and hard-normal supervision across host/transfer settings.
Panel (b) compares additional corrupted-control variants on AA-CLIP MVTec$\rightarrow$BTAD.
Values report gains over the corresponding host baseline.
}
\label{tab:cted_training_control}
\scriptsize
\setlength{\tabcolsep}{3.0pt}
\renewcommand{\arraystretch}{1.05}

\begin{minipage}[t]{0.57\textwidth}
\centering
\vspace{3pt}
\resizebox{0.9\linewidth}{!}{
\begin{tabular}{@{}llccc@{}}
\toprule
Host / transfer & Control & $\Delta$P-AUC & $\Delta$PRO & $\Delta$AP \\
\midrule
AA-CLIP MVTec$\rightarrow$BTAD & C-TED
& \textbf{+1.03} & \textbf{+3.88} & \textbf{+1.71} \\
AA-CLIP MVTec$\rightarrow$BTAD & Label swap
& -6.40 & -8.28 & -10.45 \\
\midrule
AA-CLIP VisA$\rightarrow$BTAD & C-TED
& -0.98 & \textbf{+0.69} & \textbf{+4.09} \\
AA-CLIP VisA$\rightarrow$BTAD & Label swap
& -12.19 & -15.95 & -17.23 \\
\midrule
AdaCLIP VisA$\rightarrow$BTAD & C-TED
& \textbf{+0.31} & \textbf{+2.84} & \textbf{+0.41} \\
AdaCLIP VisA$\rightarrow$BTAD & Label swap
& -0.29 & +0.45 & -0.28 \\
\bottomrule
\end{tabular}
}
\\
\textbf{(a) Transfer sanity check}
\end{minipage}
\hfill
\begin{minipage}[t]{0.40\textwidth}
\centering

\vspace{3pt}
\renewcommand{\arraystretch}{1.5}
\resizebox{\linewidth}{!}{
\begin{tabular}{@{}lccc@{}}
\toprule
Variant & $\Delta$P-AUC & $\Delta$PRO & $\Delta$AP \\
\midrule
Correct C-TED & \textbf{+1.03} & \textbf{+3.88} & \textbf{+1.71} \\
Label shuffle & -2.30 & -1.46 & -3.95 \\
Label swap & -6.40 & -8.28 & -10.45 \\
Random subspace & +0.90 & +3.34 & +0.59 \\
\bottomrule
\end{tabular}
}
\textbf{(b) Detailed controls}

\end{minipage}

\end{table*}

\subsection{Source-Supervision Semantics}
\label{subsec:supp_source_supervision}

\begin{figure}[t]
\centering
\includegraphics[width=1.0\linewidth]{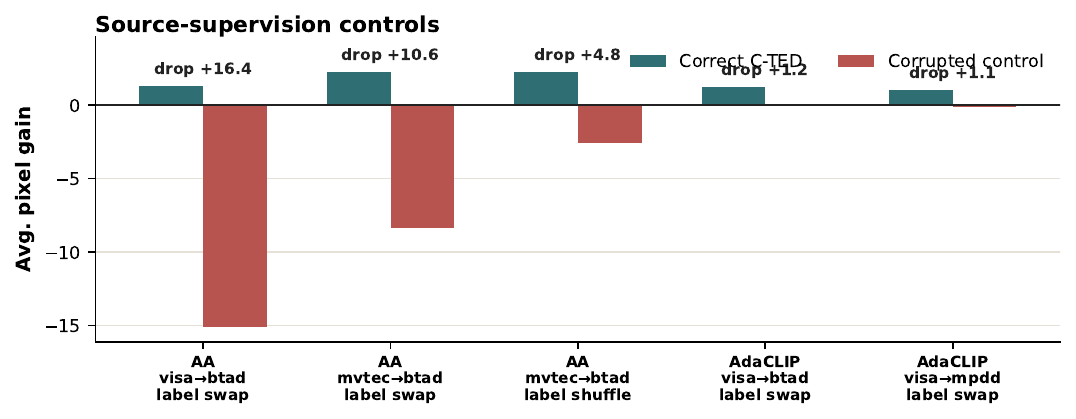}
\caption{
\textbf{Source-supervision controls for C-TED.}
All variants use the same source split, source-bank size, residual capacity, and calibration budget.
Correct C-TED preserves the defect-versus-hard-FP assignment, while label shuffle and label swap corrupt only the supervision semantics.
}
\label{fig:supp_source_supervision_control}
\end{figure}

Fig.~\ref{fig:supp_source_supervision_control} visualizes the same source-supervision control.
All variants keep the source split, source-bank size, residual capacity, and calibration budget fixed; only the defect-versus-hard-FP supervision semantics are corrupted.
The degradation under shuffled or swapped labels indicates that C-TED is not explained by adding trainable source-only residual capacity alone.

\subsection{Source Evidence in Frozen VLMs}
\label{subsec:supp_raw_vlm_support}

\begin{figure}[t]
\centering
\includegraphics[width=1.0\linewidth]{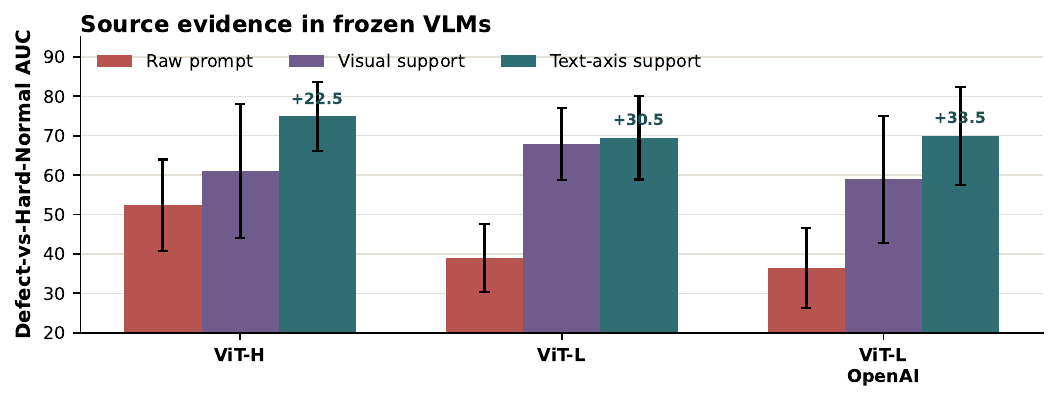}
\caption{
\textbf{Raw VLM source-support ablation.}
Raw prompt scoring is compared with source-supported readouts in frozen VLM backbones.
Text-axis support projects query and source-bank features onto the normal-versus-anomaly text axis, while visual support uses the source banks.
}
\label{fig:supp_raw_vlm_support}
\end{figure}

Fig.~\ref{fig:supp_raw_vlm_support} isolates evidence decoding before anomaly-specific host adaptation.
Raw prompt scoring often ranks hard-normal patches close to true defects.
Source-supported scoring improves defect-versus-hard-normal AUC, and the text-axis support variant is closest to \textsc{TED} because it keeps the normal-versus-anomaly text coordinate.

\subsection{Layer-Recipe Stress Test}
\label{subsec:supp_layer_stress}

Table~\ref{tab:layer_recipe_stress} tests whether layer-recipe changes alone explain the hard-FP issue.
On AdaCLIP with the full recipe, C-TED improves P-PRO from $30.48$ to $44.00$ on MPDD, with P-AP changing slightly from $29.76$ to $29.93$.
Under the last-layer recipe, AdaCLIP improves from $84.31/40.79/24.29$ to $89.62/54.00/25.95$ in P-AUC/PRO/AP.
For AnomalyCLIP, the full recipe improves from $89.94/76.56/43.52$ to $93.61/88.68/47.05$ with C-TED, while the last-layer recipe improves P-AUC from $90.39$ to $93.74$ and keeps PRO high at $88.43$.
These numbers indicate that C-TED can help after recipe changes, although the size of the effect remains host- and metric-dependent.

\begin{wraptable}{r}{0.49\textwidth}
\vspace{-2em}
\centering
\caption{
Layer-recipe stress and TED recovery on MPDD dataset.
B, T, and C denote the baseline host, train-free TED, and source-calibrated TED.
}
\label{tab:layer_recipe_stress}
\scriptsize
\setlength{\tabcolsep}{1.7pt}
\renewcommand{\arraystretch}{0.92}
\begin{tabular}{llc cccc}
\toprule
Host & Layers & M & I-AUC & P-AUC & P-PRO & P-AP \\
\midrule
\multirow{6}{*}{AdaCLIP}
& \multirow{3}{*}{full} & B & 69.01 & 96.09 & 30.48 & 29.76 \\
& & T & \textbf{69.46} & 96.08 & 30.67 & 29.70 \\
& & C & 69.25 & \textbf{96.34} & \textbf{44.00} & \textbf{29.93} \\
\cmidrule(lr){2-7}
& \multirow{3}{*}{last} & B & 61.59 & 84.31 & 40.79 & 24.29 \\
& & T & 70.00 & 84.31 & 43.06 & 24.31 \\
& & C & \textbf{70.42} & \textbf{89.62} & \textbf{54.00} & \textbf{25.95} \\
\cmidrule(lr){1-7}
\multirow{6}{*}{AnomalyCLIP}
& \multirow{3}{*}{full} & B & 93.89 & 89.94 & 76.56 & 43.52 \\
& & T & 92.51 & 93.56 & \textbf{88.73} & 46.93 \\
& & C & 93.89 & 93.61 & 88.68 & \textbf{47.05} \\
\cmidrule(lr){2-7}
& \multirow{3}{*}{last} & B & 93.89 & 90.39 & 77.82 & 43.56 \\
& & T & 90.34 & 93.66 & 88.44 & 43.73 \\
& & C & \textbf{93.91} & \textbf{93.74} & 88.43 & 43.92 \\
\bottomrule
\end{tabular}
\vspace{-2.5em}
\end{wraptable}

\subsection{Source-Bank Budget and Efficiency}
\label{subsec:supp_bank_budget_efficiency}

Table~\ref{tab:supp_efficiency_bank_budget_compact} varies the retained source-bank budget while keeping the source split, target evaluation protocol, image resolution, and post-processing fixed.
For AA-CLIP, increasing the bank budget from $64$ to $512$ changes $\Delta$Loc only from $+2.7$ to $+2.9$, with peak memory around $3.8$--$3.9$GB.
For FAPrompt, budgets from $64$ to $512$ keep $\Delta$Loc in a narrow $+8.6$--$+9.1$ range, with memory around $4.1$GB.
Runtime is not strictly monotonic across budgets, so we interpret the table as evidence of budget robustness rather than a precise scaling law.
Overall, the correction appears to saturate with a moderate retained bank budget, supporting a fixed source-bank policy without target-domain score selection.

\begin{table}[t]
\centering
\caption{Fixed source-bank budget sensitivity. Values are averaged over completed transfer settings for each host and bank budget. $\Delta$Loc is the mean of $\Delta$PRO and $\Delta$AP over the host baseline.}
\label{tab:supp_efficiency_bank_budget_compact}

\setlength{\tabcolsep}{3.6pt}
\renewcommand{\arraystretch}{1.08}
\resizebox{0.6\linewidth}{!}{%
\begin{tabular}{lrrrrrr}
\toprule
Host & Budget & $n$ & Retained & ms/img & Mem. & $\Delta$Loc \\
\midrule
AA-CLIP & 64 & 6 & 6634.0 & 206.8 & 3.8GB & 2.7 \\
AA-CLIP & 128 & 6 & 12180.0 & 182.9 & 3.8GB & 2.9 \\
AA-CLIP & 512 & 6 & 35238.0 & 149.3 & 3.9GB & 2.9 \\
FAPrompt & 64 & 6 & 256.0 & 1070.1 & 4.1GB & 9.1 \\
FAPrompt & 128 & 6 & 512.0 & 1041.9 & 4.1GB & 8.6 \\
FAPrompt & 256 & 6 & 1024.0 & 1055.2 & 4.1GB & 9.1 \\
FAPrompt & 512 & 6 & 2048.0 & 1022.8 & 4.1GB & 9.1 \\
\bottomrule
\end{tabular}%
}
\end{table}

\subsection{Full-Feature Memory Replacement Control}
\label{subsec:supp_fullfeature_memory_control}
Table~\ref{tab:supp_fullfeature_memory_control_full} provides transfer-wise full-feature memory replacement controls.
This control uses the same source visual banks but removes the text-conditioned evidence readout, replacing the local map with a full-feature defect-versus-hard-FP memory score.
The result is mostly negative.
For FAPrompt, all listed transfers degrade, for example MVTec$\rightarrow$BTAD changes by $-44.60/-47.22/-15.91$ in P-AUC/PRO/AP, and MVTec$\rightarrow$VisA changes by $-36.61/-59.25/-8.92$.
For AdaptCLIP, degradation is also large, including MVTec$\rightarrow$MPDD at $-68.89/-87.24/-23.90$ and MVTec$\rightarrow$VisA at $-63.02/-84.15/-25.80$.
AA-CLIP contains one positive case, MVTec$\rightarrow$MPDD at $+0.58/+4.92/+3.27$, but most AA-CLIP transfers are still negative.
This supports a cautious interpretation: generic visual memory retrieval is not a substitute for the text-conditioned defect-versus-hard-FP comparison used by \textsc{TED}.
\paragraph{Extended same-bank retrieval controls.}
We extend the full-feature comparison with 1-NN, prototype,
and LogMeanExp readouts, each evaluated in replacement and
residual modes.
The comparison covers all 18 settings across AA-CLIP,
AdaCLIP, and FAPrompt, with six transfers per host.
Source banks, hosts, targets, and aggregation are held fixed;
only the evidence readout changes.

As shown in Table~\ref{tab:same_bank_controls},
none of the tested Full-D readouts improves all three
pixel-level metrics on average, whereas C-TED does.
The prototype residual achieves a larger P-PRO gain,
but reduces P-AUC and P-AP.
These results indicate that access to source banks or
generic full-dimensional similarity alone does not explain
the joint improvements observed with C-TED.

\begin{table}[t]
\centering
\caption{
Transfer-wise full-feature memory replacement control.
The control uses the same source visual banks but removes the text-conditioned readout.
Values are gains over the corresponding host baseline.
}
\label{tab:supp_fullfeature_memory_control_full}
\scriptsize
\setlength{\tabcolsep}{3.0pt}
\renewcommand{\arraystretch}{0.98}
\begin{tabular}{@{}llrrr@{}}
\toprule
Host & Transfer & $\Delta$P-AUC & $\Delta$PRO & $\Delta$AP \\
\midrule
\multirow{6}{*}{AA-CLIP}
& MVTec$\to$BTAD & -5.75 & -7.53 & -11.35 \\
& MVTec$\to$MPDD & +0.58 & +4.92 & +3.27 \\
& MVTec$\to$VisA & -0.43 & -1.58 & -3.28 \\
& VisA$\to$BTAD & -13.48 & -16.22 & -27.88 \\
& VisA$\to$MPDD & -2.53 & -6.18 & -5.98 \\
& VisA$\to$MVTec & -5.45 & -6.30 & -12.33 \\
\midrule
\multirow{6}{*}{FAPrompt}
& MVTec$\to$BTAD & -44.60 & -47.22 & -15.91 \\
& MVTec$\to$MPDD & -37.18 & -52.31 & -15.68 \\
& MVTec$\to$VisA & -36.61 & -59.25 & -8.92 \\
& VisA$\to$BTAD & -51.69 & -48.15 & -15.30 \\
& VisA$\to$MPDD & -38.00 & -44.62 & -14.72 \\
& VisA$\to$MVTec & -40.32 & -50.61 & -10.58 \\
\midrule
\multirow{6}{*}{AdaptCLIP}
& MVTec$\to$BTAD & -45.82 & -55.90 & -39.55 \\
& MVTec$\to$MPDD & -68.89 & -87.24 & -23.90 \\
& MVTec$\to$VisA & -63.02 & -84.15 & -25.80 \\
& VisA$\to$BTAD & -41.15 & -61.49 & -43.17 \\
& VisA$\to$MPDD & -24.01 & -45.58 & -21.96 \\
& VisA$\to$MVTec & -25.35 & -37.33 & -26.88 \\
\bottomrule
\end{tabular}
\end{table}

\begin{table}[t]
  \centering
  \caption{
    Extended same-bank retrieval controls across AA-CLIP,
    AdaCLIP, and FAPrompt, with six transfers per host
    (18 settings in total).
    Values are mean changes in percentage points relative
    to the corresponding host.
    Source banks, hosts, targets, and aggregation are fixed;
    only the evidence readout changes.
    Replacement substitutes the host local score, whereas
    residual mode adds a correction to it.
  }
  \label{tab:same_bank_controls}
  \small
  \setlength{\tabcolsep}{7pt}
  \begin{tabular}{llrrr}
    \toprule
    Readout & Mode
      & $\Delta$P-AUC & $\Delta$P-PRO & $\Delta$P-AP \\
    \midrule
    Full-D 1-NN
      & Replacement & $-45.65$ & $-39.39$ & $-20.68$ \\
      & Residual    & $-10.99$ & $-0.22$  & $-7.66$  \\
    Full-D prototype
      & Replacement & $-11.85$ & $-3.13$  & $-12.63$ \\
      & Residual    & $-3.27$  & $+11.04$ & $-5.64$  \\
    Full-D LogMeanExp
      & Replacement & $-24.70$ & $-18.56$ & $-16.67$ \\
      & Residual    & $-8.68$  & $+3.69$  & $-7.86$  \\
    \midrule
    C-TED
      & Residual    & $+0.87$  & $+4.38$  & $+5.21$  \\
    \bottomrule
  \end{tabular}
\end{table}




\begin{table}[t]
\centering
\caption{
Weak-source robustness.
AA-CLIP varies the number of source images per source class used to construct banks.
FAPrompt varies the retained source-bank budget per class.
Values are gains over the corresponding host baseline; $\Delta$Loc is the mean of $\Delta$PRO and $\Delta$AP.
}
\label{tab:weak_source_robustness}
\scriptsize
\setlength{\tabcolsep}{3.2pt}
\renewcommand{\arraystretch}{1.06}
\begin{tabular*}{\linewidth}{@{\extracolsep{\fill}}llcccc}
\toprule
Host & Source budget & $n$ & $\Delta$PRO & $\Delta$AP & $\Delta$Loc \\
\midrule
AA-CLIP & 1 img/cls & 4 & +3.3 & +1.9 & +2.6 \\
AA-CLIP & 2 img/cls & 4 & +3.5 & +1.7 & +2.6 \\
AA-CLIP & 4 img/cls & 4 & +3.2 & +3.2 & +3.2 \\
AA-CLIP & all img/cls & 4 & +3.1 & +2.9 & +3.0 \\
FAPrompt & 8 bank/cls & 3 & +7.2 & +13.4 & +10.3 \\
FAPrompt & 16 bank/cls & 3 & +7.3 & +13.4 & +10.4 \\
FAPrompt & 32 bank/cls & 3 & +7.2 & +13.3 & +10.3 \\
FAPrompt & 64 bank/cls & 3 & +7.2 & +13.3 & +10.3 \\
\bottomrule
\end{tabular*}
\end{table}

\subsection{Weak-Source and Bank-Budget Robustness}

Table~\ref{tab:weak_source_robustness} evaluates the sensitivity of TED to reduced source evidence.
For AA-CLIP, we vary the number of source images per class used to construct the evidence banks.
For FAPrompt, we vary the retained source-bank budget per class after bank construction.
The source split, target evaluation protocol, residual rank, insertion strength, post-processing, and target-domain selection policy are kept fixed.

The gains remain positive in these reduced-source settings.
For AA-CLIP, even the smallest tested source-image budget retains positive localization gains.
For FAPrompt, reducing the retained bank budget from 64 to 8 entries per class gives similar localization gains in the completed settings.
These results suggest that the observed improvements are not driven only by retaining a very large source bank.
At the same time, this is not a zero-source setting: representative source defect and hard-FP evidence remains necessary for constructing the comparison.
\subsection{Simple Source-Supervised Readout Controls}
\label{subsec:supp_source_readout_controls}

\begin{table}[t]
\centering
\caption{
Source-supervised readout control.
The control uses the same source defect and hard-FP banks as C-TED, but replaces the evidence comparison with a directly learned MLP rank readout.
Replace uses the learned readout as the local anomaly map, while Residual adds it to the host map.
Rows average completed transfer settings for each host.
Values are $\Delta$Loc, the mean of $\Delta$PRO and $\Delta$AP over the host baseline.
}
\label{tab:supp_source_readout_extra_controls}
\scriptsize
\setlength{\tabcolsep}{3.0pt}
\renewcommand{\arraystretch}{1.05}
\begin{tabular*}{\linewidth}{@{\extracolsep{\fill}}lcccc}
\toprule
Host & $n$ & Replace & Residual & C-TED \\
\midrule
AA-CLIP & 4 & -7.4 & -4.7 & \textbf{+3.0} \\
AdaCLIP & 4 & -12.3 & -4.1 & \textbf{+4.1} \\
FAPrompt & 4 & -42.9 & +3.8 & \textbf{+10.0} \\

\bottomrule
\end{tabular*}
\end{table}

We compare C-TED with a simple source-supervised MLP rank readout trained from the same source defect and hard-FP banks.
This control uses the same source split and target evaluation protocol as C-TED, but replaces the TED evidence comparison with a direct learned scalar readout.
We evaluate two usage modes: replacing the local anomaly map with the learned readout, and adding the learned readout as a residual to the host map.

Table~\ref{tab:supp_source_readout_extra_controls} shows that direct replacement is brittle because it discards the host-specific local calibration.
Adding the learned readout as a residual is more stable, but remains weaker than C-TED in the completed MLP-rank settings.
This suggests that the gains are not explained only by access to the same source labels and source banks.
Rather, source evidence is more effective when used as a defect-versus-hard-normal comparison anchored to the host response, instead of as a generic learned scalar readout.

\subsection{Residual-Strength Boundary}
\label{subsec:supp_residual_strength_boundary}

We use a small strength-sweep diagnostic to check whether the source-driven residual should simply be made larger.
For this diagnostic, we keep the source banks, source calibrator, residual basis, target evaluation protocol, image resolution, and post-processing fixed, and vary only the residual insertion strength $\alpha$.
Table~\ref{tab:supp_residual_strength_boundary} reports the resulting localization gain on AA-CLIP.

\begin{wraptable}{r}{0.39\linewidth}
\centering
\vspace{-0.7em}
\caption{
\textbf{Residual-strength boundary.}
AA-CLIP diagnostic with fixed calibrator and protocol.
}
\label{tab:supp_residual_strength_boundary}
{\fontsize{5.8pt}{6.3pt}\selectfont
\setlength{\tabcolsep}{1.8pt}
\renewcommand{\arraystretch}{0.82}
\resizebox{0.94\linewidth}{!}{%
\begin{tabular}{@{}lccccc@{}}
\toprule
$\alpha$ & 0 & 0.25 & 0.5 & 1 & 2 \\
\midrule
$\Delta$Loc & +0.0 & \textbf{+0.3} & +0.3 & -0.0 & -5.2 \\
\bottomrule
\end{tabular}%
}
}
\vspace{-1.0em}
\end{wraptable}

The sweep suggests that stronger insertion is not automatically better.
Moderate residual strengths give small positive gains, while an overly strong residual can reduce localization quality.
We use this result as a boundary diagnostic rather than as a universal strength-sweep claim.
It supports the design choice of keeping C-TED as a bounded correction anchored to the host response, instead of allowing the source-driven residual to dominate the host map.

\subsection{Host-Score-Only Calibration Control}
\label{subsec:supp_host_score_only_control}

We additionally test whether C-TED can be reduced to a scalar recalibration of the host local score.
This control uses the same source defect and hard-FP patch sets as C-TED, but removes the source support features.
It trains a source-only rank readout from only the scalar host scores observed on source defect and hard-FP patches.
We evaluate two variants: replacing the local anomaly map with the learned readout, and adding the learned readout as a residual to the host map.

\begin{table}[t]
\centering
\caption{
Host-score-only calibration control.
Controls train a source-only rank readout using only the scalar host local score on source defect and hard-FP patches, without source defect/hard-FP support
features.
Each entry reports $\Delta$PRO / $\Delta$AP over the corresponding host baseline.
}
\label{tab:supp_host_score_only_control}
\scriptsize
\setlength{\tabcolsep}{3.0pt}
\renewcommand{\arraystretch}{1.05}
\begin{tabular*}{\linewidth}{@{\extracolsep{\fill}}lcccc}
\toprule
Host & $n$ & Host replace & Host residual & C-TED \\
\midrule
AA-CLIP & 4
& $+0.1 / -4.5$
& $-2.3 / -7.7$
& $\mathbf{+3.1 / +2.9}$ \\
FAPrompt & 4
& $+3.0 / +3.3$
& $+4.3 / +5.0$
& $\mathbf{+6.4 / +13.5}$ \\
AdaCLIP & 4
& $+5.3 / -9.7$
& $+0.1 / -5.9$
& $\mathbf{+7.4 / -0.3}$ \\

\bottomrule
\end{tabular*}
\end{table}

Table~\ref{tab:supp_host_score_only_control} shows that host-score-only calibration is not a reliable substitute for C-TED.
The scalar readout can improve one metric in some hosts, but it often reduces AP or weakens the overall correction.
For AA-CLIP, both host-score-only variants decrease AP, whereas C-TED improves both PRO and AP.
For FAPrompt and AdaCLIP, C-TED also gives the strongest AP change among the compared variants.
These results suggest that C-TED is not only recalibrating the host score; the source defect and hard-FP support features provide additional evidence for
resolving ambiguous local responses.
\section{Additional Qualitative and Diagnostic Results}
\label{sec:supp_qualitative_diagnostics}

\subsection{Raw VLM Failure and Recovery}
\label{subsec:supp_raw_vlm_failure_recovery}

\begin{figure}[t]
\centering
\includegraphics[width=1.0\linewidth]{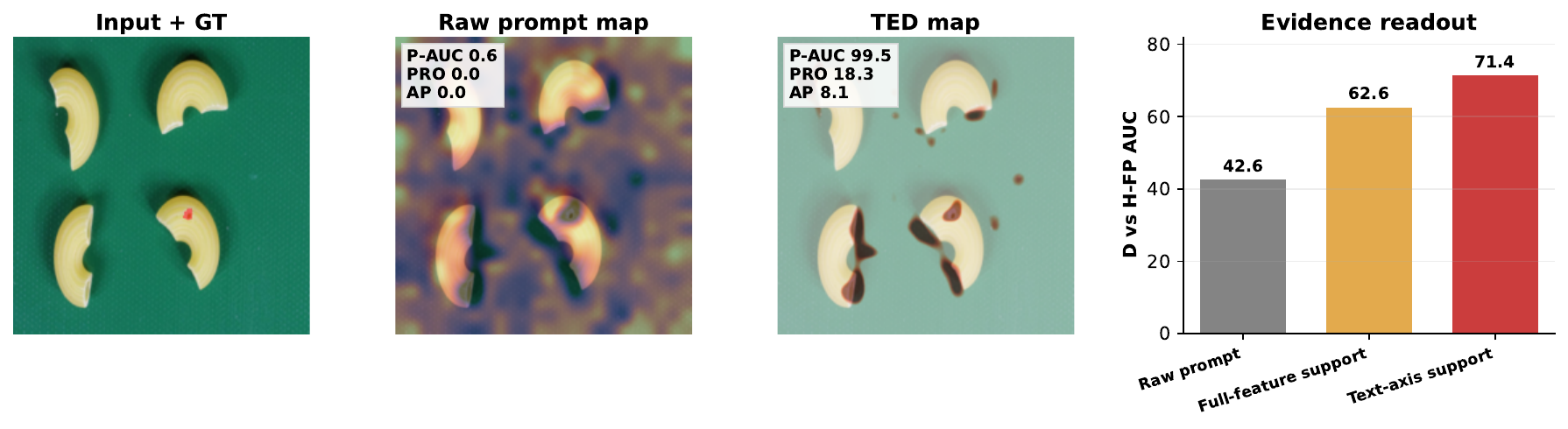}
\caption{
\textbf{Raw VLM failure and TED recovery.}
Raw prompt scoring confuses defect evidence with hard-normal saliency, while TED recovers a cleaner local ranking using source-supported evidence.
}
\label{fig:supp_raw_vlm_clean_heatmap_recovery}
\end{figure}

Fig.~\ref{fig:supp_raw_vlm_clean_heatmap_recovery} illustrates why raw prompt scoring fails in frozen CLIP.
The raw prompt map responds strongly to visually salient normal structures and does not cleanly isolate the defect.
\textsc{TED} keeps the same frozen backbone and prompts, but re-ranks the local response using source defect and hard-FP support.

\subsection{Failure-Conditioned Separability Robustness}
\label{subsec:supp_failure_conditioned_separability}

Table~\ref{tab:failure_tab} expands the failure-conditioned diagnostic to 70 completed adapted-host settings.
For each setting, hard-normal regions are selected using only the baseline response, and we compare how the baseline and TED rank anomalous regions relative to these hard-normal competitors.
Across the completed settings, TED improves the AUC for separating anomalous and hard-normal regions, increases the mean score margin between the two groups, and improves localization metrics with
positive setting-level bootstrap intervals.
This shows that the localization gains are accompanied by better separation of anomalous regions from hard-normal competitors, rather than only by a uniform shift of anomaly scores.
We therefore use this analysis as mechanism evidence that TED reduces hard-normal competition, while not interpreting baseline severity as a universal linear predictor of localization gain for every host.
\subsection{Qualitative Localization on an Adapted Host}
\label{app:adapted_host_qualitative}

\begin{figure}[p]
  \centering
  \includegraphics[
    width=\linewidth,
    height=0.80\textheight,
    keepaspectratio
  ]{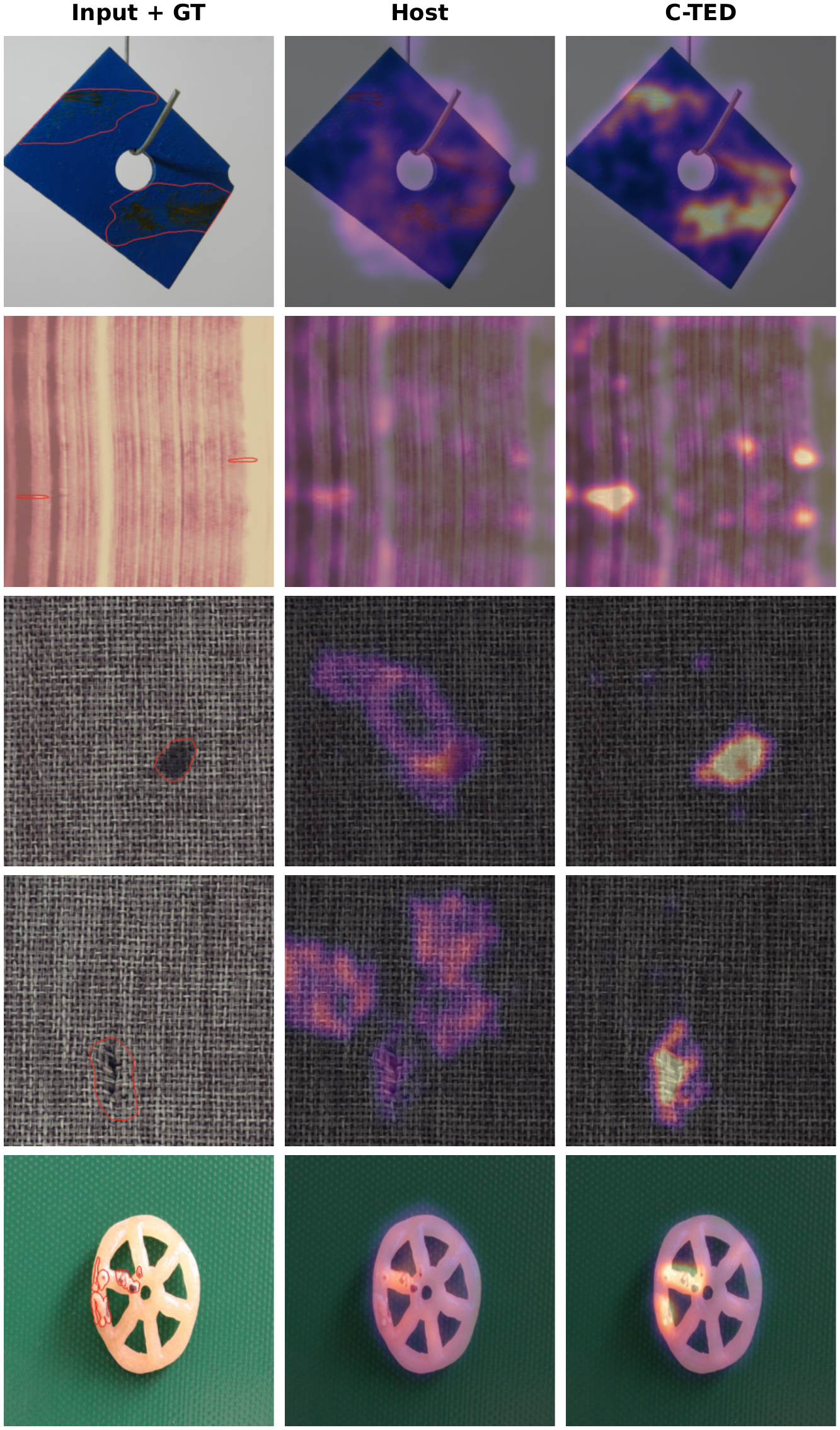}
\caption{Qualitative comparisons of FAPrompt and C-TED across four target datasets, with ground-truth defects outlined in red and a shared color scale for each Host/C-TED pair.}
  \label{fig:supp_cted_qualitative}
\end{figure}

Figure~\ref{fig:supp_cted_qualitative} presents selected
qualitative examples of C-TED on FAPrompt across four
target datasets.

\begin{table}[t]
\centering
\caption{
Failure-conditioned separability robustness.
Intervals are setting-level percentile bootstrap 95\% CIs over completed adapted-host settings.
$\Delta$AUC measures the change in AUC for ranking anomalous regions above hard normal regions.
$\Delta$Loc is the mean of $\Delta$PRO and $\Delta$AP.
}
\label{tab:failure_tab}
\footnotesize
\setlength{\tabcolsep}{4.0pt}
\renewcommand{\arraystretch}{1.08}
\begin{tabular*}{\linewidth}{@{\extracolsep{\fill}}lccc@{}}
\toprule
Scope & $n$ & $\Delta$AUC [95\% CI] & $\Delta$Loc [95\% CI] \\
\midrule
FAPrompt & 36 & +8.1 [+6.3, +10.0] & +8.4 [+7.1, +9.7] \\
AA-CLIP & 10 & +2.2 [-1.9, +5.9] & +11.3 [-2.0, +25.1] \\
AdaCLIP & 12 & -2.5 [-4.3, -1.1] & +4.6 [+3.5, +5.7] \\
AdaptCLIP & 12 & +1.2 [+1.1, +1.3] & +0.3 [+0.1, +0.5] \\
All adapted hosts & 70 & +4.3 [+2.8, +5.8] & +6.8 [+4.7, +9.1] \\
\bottomrule
\end{tabular*}
\end{table}

\begin{figure}[t]
\centering
\includegraphics[width=1.0\linewidth]{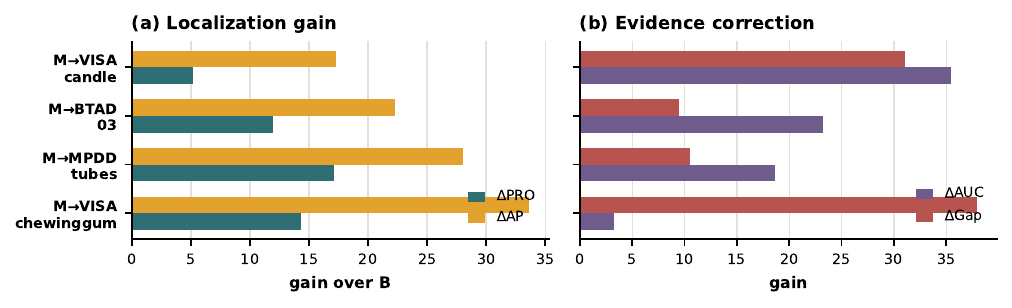}
\caption{
\textbf{Failure-conditioned class-level cases.}
Representative target classes where C-TED improves localization while also increasing defect-versus-hard-FP AUC and the defect-minus-hard-FP score gap.
}
\label{fig:direct_failure_conditioned_cases}
\end{figure}

\section{Additional Quantitative Results}
\label{sec:supp_quantitative}

\subsection{Full Adapted-Host Results}
\label{subsec:supp_adapted_hosts}

The following tables provide detailed transfer results for adapted CLIP-AD hosts.
They complement the main-paper summaries by showing the behavior across host architectures, visual backbones, target datasets, and metrics.
We interpret these tables as detailed evidence for a local evidence-decoding effect, not as a claim that every host, backbone, and metric improves uniformly.
The clearest gains appear when the host still ranks hard-normal evidence close to true defects, while already calibrated or saturated settings can show smaller or mixed changes.

\subsubsection{AA-CLIP}
\label{subsec:supp_aaclip}

Table~\ref{tab:aaclip_hostres_only} reports AA-CLIP with host-anchored residual correction.
Several settings show large pixel-level recovery from weak baseline localization.
For ViT-B/16+ on MVTec, P-AUC/PRO/AP improve from $39.27/11.86/2.97$ to $69.22/38.88/10.39$.
On ViT-B/16+ BTAD, the same metrics improve from $61.37/30.66/8.51$ to $76.30/41.13/13.73$.
For ViT-H/14 on MVTec, C-TED improves P-AUC/PRO/AP from $51.04/16.10/4.36$ to $64.60/26.06/5.65$.
These cases suggest that C-TED can recover useful local ranking when the baseline anomaly map is weak and hard-normal responses remain competitive.
At the same time, the table includes boundary cases.
For example, ViT-L/14-336 on BTAD has lower PRO/AP after correction despite a strong baseline.
This pattern is consistent with our main claim: \textsc{TED} is most useful when there is hard-FP competition to correct, and more conservative or mixed when the host is already well calibrated.

\begin{table*}[h]
\centering
\caption{AA-CLIP with host-anchored residual subspace correction. Results are averaged over three seeds; each entry is mean{\scriptsize $\pm$std}.}
\label{tab:aaclip_hostres_only}
\small
\setlength{\tabcolsep}{4pt}
\renewcommand{\arraystretch}{1.08}
\resizebox{\textwidth}{!}{
\begin{tabular}{lllc cccc}
\toprule
Host & Backbone & Target & Method & I-AUROC & P-AUROC & P-PRO & P-AP \\
\midrule

\multirow{32}{*}{\begin{tabular}[c]{@{}c@{}}AA-CLIP\end{tabular}}

& \multirow{8}{*}{ViT-B/16+}
& \multirow{2}{*}{MVTec} & Base & 50.81$\pm$0.00 & 39.27$\pm$0.00 & 11.86$\pm$0.00 & 2.97$\pm$0.00 \\
& & & OURS & \textbf{60.67$\pm$0.05} & \textbf{69.22$\pm$0.35} & \textbf{38.88$\pm$0.82} & \textbf{10.39$\pm$0.17} \\
\cmidrule(lr){3-8}

& & \multirow{2}{*}{VisA} & Base & 46.61$\pm$0.00 & 54.49$\pm$0.00 & 22.32$\pm$0.00 & 1.31$\pm$0.00 \\
& & & OURS & \textbf{54.41$\pm$0.14} & \textbf{63.66$\pm$0.22} & \textbf{28.74$\pm$0.29} & \textbf{2.18$\pm$0.04} \\
\cmidrule(lr){3-8}

& & \multirow{2}{*}{MPDD} & Base & {44.86$\pm$0.00} & {88.93$\pm$0.00} & {65.37$\pm$0.00} & \textbf{12.10$\pm$0.00} \\
& & & OURS & \textbf{48.04$\pm$0.21} & \textbf{88.98$\pm$0.28} & \textbf{65.44$\pm$0.67} & 12.07$\pm$0.05 \\
\cmidrule(lr){3-8}

& & \multirow{2}{*}{BTAD} & Base & 63.83$\pm$0.00 & 61.37$\pm$0.00 & 30.66$\pm$0.00 & 8.51$\pm$0.00 \\
& & & OURS & \textbf{79.78$\pm$0.07} & \textbf{76.30$\pm$0.11} & \textbf{41.13$\pm$0.08} & \textbf{13.73$\pm$0.11} \\

\cmidrule(lr){2-8}

& \multirow{8}{*}{ViT-L/14 OpenAI}
& \multirow{2}{*}{MVTec} & Base & 84.80$\pm$0.00 & \textbf{90.22$\pm$0.00} & 78.94$\pm$0.00 & 34.51$\pm$0.00 \\
& & & OURS & \textbf{85.19$\pm$0.00} & 89.91$\pm$0.00 & \textbf{81.45$\pm$0.08} & \textbf{35.80$\pm$0.02} \\
\cmidrule(lr){3-8}

& & \multirow{2}{*}{VisA} & Base & 72.04$\pm$0.00 & 92.27$\pm$0.00 & 73.85$\pm$0.00 & 17.69$\pm$0.00 \\
& & & OURS & \textbf{73.56$\pm$0.00} & \textbf{93.50$\pm$0.01} & \textbf{75.07$\pm$0.05} & \textbf{20.57$\pm$0.01} \\
\cmidrule(lr){3-8}

& & \multirow{2}{*}{MPDD} & Base & 67.67$\pm$0.00 & 94.15$\pm$0.00 & 78.51$\pm$0.00 & 17.30$\pm$0.00 \\
& & & OURS & \textbf{71.50$\pm$0.00} & \textbf{95.25$\pm$0.00} & \textbf{83.33$\pm$0.04} & \textbf{23.01$\pm$0.02} \\
\cmidrule(lr){3-8}

& & \multirow{2}{*}{BTAD} & Base & 90.62$\pm$0.00 & 92.48$\pm$0.00 & 59.78$\pm$0.00 & 36.55$\pm$0.00 \\
& & & OURS & \textbf{91.46$\pm$0.00} & \textbf{93.50$\pm$0.01} & \textbf{63.56$\pm$0.09} & \textbf{38.25$\pm$0.01} \\

\cmidrule(lr){2-8}

& \multirow{8}{*}{ViT-L/14-336}
& \multirow{2}{*}{MVTec} & Base & \textbf{89.78$\pm$0.00} & \textbf{91.09$\pm$0.00} & 83.16$\pm$0.00 & \textbf{43.57$\pm$0.00} \\
& & & OURS & 89.42$\pm$0.00 & 90.25$\pm$0.01 & \textbf{83.31$\pm$0.08} & 41.66$\pm$0.02 \\
\cmidrule(lr){3-8}

& & \multirow{2}{*}{VisA} & Base & 74.26$\pm$0.00 & \textbf{94.26$\pm$0.00} & \textbf{79.64$\pm$0.00} & \textbf{21.75$\pm$0.00} \\
& & & OURS & \textbf{75.22$\pm$0.01} & 94.06$\pm$0.00 & 79.15$\pm$0.17 & 21.40$\pm$0.02 \\
\cmidrule(lr){3-8}

& & \multirow{2}{*}{MPDD} & Base & 58.48$\pm$0.00 & 96.05$\pm$0.00 & 85.66$\pm$0.00 & 21.60$\pm$0.00 \\
& & & OURS & \textbf{67.88$\pm$0.04} & \textbf{96.08$\pm$0.00} & \textbf{85.94$\pm$0.01} & \textbf{22.68$\pm$0.02} \\
\cmidrule(lr){3-8}

& & \multirow{2}{*}{BTAD} & Base & \textbf{94.24$\pm$0.00} & \textbf{93.59$\pm$0.00} & \textbf{67.70$\pm$0.00} & \textbf{39.64$\pm$0.00} \\
& & & OURS & 94.19$\pm$0.00 & 90.68$\pm$0.01 & 57.67$\pm$0.00 & 32.53$\pm$0.04 \\

\cmidrule(lr){2-8}

& \multirow{8}{*}{ViT-H/14}
& \multirow{2}{*}{MVTec} & Base & 46.46$\pm$1.80 & 51.04$\pm$1.85 & 16.10$\pm$0.63 & 4.36$\pm$0.03 \\
& & & OURS & \textbf{53.25$\pm$1.78} & \textbf{72.61$\pm$4.21} & \textbf{40.36$\pm$8.33} & \textbf{9.10$\pm$2.86} \\
\cmidrule(lr){3-8}

& & \multirow{2}{*}{VisA} & Base & 47.77$\pm$1.53 & 41.62$\pm$7.06 & 13.42$\pm$4.78 & 0.83$\pm$0.19 \\
& & & OURS & \textbf{49.58$\pm$1.82} & \textbf{77.24$\pm$0.75} & \textbf{47.56$\pm$0.93} & \textbf{2.57$\pm$0.30} \\
\cmidrule(lr){3-8}

& & \multirow{2}{*}{MPDD} & Base & \textbf{51.81$\pm$1.13} & 45.68$\pm$2.88 & 14.95$\pm$2.64 & \textbf{5.83$\pm$2.78} \\
& & & OURS & 50.93$\pm$0.19 & \textbf{68.93$\pm$4.34} & \textbf{28.86$\pm$5.12} & 3.79$\pm$0.58 \\
\cmidrule(lr){3-8}

& & \multirow{2}{*}{BTAD} & Base & 53.34$\pm$4.58 & 35.25$\pm$4.11 & 8.04$\pm$1.22 & 2.26$\pm$0.33 \\
& & & OURS & \textbf{62.19$\pm$8.35} & \textbf{76.31$\pm$3.33} & \textbf{44.24$\pm$2.62} & \textbf{11.11$\pm$2.83} \\

\bottomrule
\end{tabular}
}
\end{table*}




\subsubsection{FAPrompt}
\label{subsec:supp_faprompt}

Table~\ref{tab:faprompt_results} reports FAPrompt results.
C-TED gives particularly clear gains on BTAD, where hard normal structures often compete with subtle defect evidence.
For ViT-L/14-336 on BTAD, P-AUC/PRO/AP improve from $90.52/60.32/21.54$ to $94.99/70.95/45.01$.
The AP gain is substantial, suggesting that the correction improves sparse defect localization rather than only shifting broad pixel scores.
For ViT-L/14 OpenAI on BTAD, AP improves from $16.51$ to $40.96$, while P-AUC/PRO improve from $87.19/54.00$ to $93.71/59.99$.
These two BTAD cases are useful because the gains appear not only in ranking-based P-AUC, but also in PRO and AP, indicating improved region-level and sparse-pixel localization.
This is consistent with FAPrompt already making the host defect-sensitive, while C-TED further separates hard-normal-supported responses from defect-supported responses.
Some settings show smaller gains, so we treat FAPrompt as a strong but still host- and dataset-dependent case.

\begin{table*}[h]
\centering
\caption{Results with \textbf{FAPrompt} as host.}
\label{tab:faprompt_results}
\small
\setlength{\tabcolsep}{4pt}
\renewcommand{\arraystretch}{1.08}
\resizebox{\textwidth}{!}{
\begin{tabular}{lllc cccc}
\toprule
Host & Backbone & Target & Method & I-AUROC & P-AUROC & P-PRO & P-AP \\
\midrule

\multirow{32}{*}{\begin{tabular}[c]{@{}c@{}}FAPrompt\end{tabular}}

& \multirow{8}{*}{ViT-B/16+}
& \multirow{2}{*}{MVTec} & Base & 80.75$\pm$0.00 & 65.61$\pm$0.00 & 35.51$\pm$0.00 & 8.78$\pm$0.00 \\
& & & OURS & 80.75$\pm$0.00 & \textbf{77.95$\pm$0.00} & \textbf{62.34$\pm$0.01} & \textbf{24.52$\pm$0.00} \\
\cmidrule(lr){3-8}

& & \multirow{2}{*}{VisA} & Base & 69.16$\pm$0.00 & 67.98$\pm$0.00 & 40.10$\pm$0.00 & 3.61$\pm$0.00 \\
& & & OURS & 69.16$\pm$0.00 & \textbf{71.56$\pm$0.01} & \textbf{54.98$\pm$0.08} & \textbf{8.79$\pm$0.00} \\
\cmidrule(lr){3-8}

& & \multirow{2}{*}{MPDD} & Base & 67.35$\pm$0.00 & 81.33$\pm$0.00 & 59.09$\pm$0.00 & 11.07$\pm$0.00 \\
& & & OURS & 67.35$\pm$0.00 & \textbf{82.45$\pm$0.01} & \textbf{61.23$\pm$0.17} & \textbf{12.98$\pm$0.01} \\
\cmidrule(lr){3-8}

& & \multirow{2}{*}{BTAD} & Base & 77.86$\pm$0.00 & 72.46$\pm$0.00 & 34.31$\pm$0.00 & 5.92$\pm$0.00 \\
& & & OURS & 77.86$\pm$0.00 & \textbf{79.83$\pm$0.01} & \textbf{47.19$\pm$0.06} & \textbf{19.62$\pm$0.01} \\

\cmidrule(lr){2-8}

& \multirow{8}{*}{ViT-L/14 OpenAI}
& \multirow{2}{*}{MVTec} & Base & 85.42$\pm$0.00 & 88.56$\pm$0.00 & 68.59$\pm$0.00 & 18.58$\pm$0.00 \\
& & & OURS & 85.42$\pm$0.00 & \textbf{89.80$\pm$0.00} & \textbf{74.45$\pm$0.00} & \textbf{33.34$\pm$0.00} \\
\cmidrule(lr){3-8}
& & \multirow{2}{*}{VisA} & Base & 80.32$\pm$0.00 & 93.60$\pm$0.00 & 74.68$\pm$0.00 & 10.73$\pm$0.00 \\
& & & OURS & 80.32$\pm$0.00 & \textbf{95.12$\pm$0.00} & \textbf{83.90$\pm$0.00} & \textbf{17.89$\pm$0.00} \\
\cmidrule(lr){3-8}
& & \multirow{2}{*}{MPDD} & Base & 80.01$\pm$0.00 & 93.10$\pm$0.00 & 74.43$\pm$0.00 & 13.42$\pm$0.00 \\
& & & OURS & 80.01$\pm$0.00 & \textbf{94.81$\pm$0.00} & \textbf{80.85$\pm$0.00} & \textbf{19.69$\pm$0.00} \\
\cmidrule(lr){3-8}
& & \multirow{2}{*}{BTAD} & Base & 86.10$\pm$0.00 & 87.19$\pm$0.00 & 54.00$\pm$0.00 & 16.51$\pm$0.00 \\
& & & OURS & 86.10$\pm$0.00 & \textbf{93.71$\pm$0.00} & \textbf{59.99$\pm$0.00} & \textbf{40.96$\pm$0.00} \\
\cmidrule(lr){2-8}

& \multirow{8}{*}{ViT-L/14-336}
& \multirow{2}{*}{MVTec} & Base & 91.47$\pm$0.00 & 88.88$\pm$0.00 & 70.72$\pm$0.00 & 16.92$\pm$0.00 \\
& & & OURS & 91.47$\pm$0.00 & \textbf{90.29$\pm$0.00} & \textbf{75.33$\pm$0.00} & \textbf{31.00$\pm$0.00} \\
\cmidrule(lr){3-8}
& & \multirow{2}{*}{VisA} & Base & 82.76$\pm$0.00 & 94.42$\pm$0.00 & 77.10$\pm$0.00 & 10.43$\pm$0.00 \\
& & & OURS & 82.76$\pm$0.00 & \textbf{95.61$\pm$0.00} & \textbf{82.39$\pm$0.00} & \textbf{19.21$\pm$0.00} \\
\cmidrule(lr){3-8}
& & \multirow{2}{*}{MPDD} & Base & 79.45$\pm$0.00 & 94.17$\pm$0.00 & 78.85$\pm$0.00 & 17.59$\pm$0.00 \\
& & & OURS & 79.45$\pm$0.00 & \textbf{96.07$\pm$0.00} & \textbf{83.91$\pm$0.00} & \textbf{25.31$\pm$0.00} \\
\cmidrule(lr){3-8}
& & \multirow{2}{*}{BTAD} & Base & 89.24$\pm$0.00 & 90.52$\pm$0.00 & 60.32$\pm$0.00 & 21.54$\pm$0.00 \\
& & & OURS & 89.24$\pm$0.00 & \textbf{94.99$\pm$0.00} & \textbf{70.95$\pm$0.00} & \textbf{45.01$\pm$0.00} \\

\cmidrule(lr){2-8}

& \multirow{8}{*}{ViT-H/14}
& \multirow{2}{*}{MVTec} & Base & 71.22 & 75.62 & 44.23 & 10.50 \\
& & & OURS & 71.22 & \textbf{77.23} & \textbf{53.27} & \textbf{18.87} \\
\cmidrule(lr){3-8}

& & \multirow{2}{*}{VisA} & Base & 61.05 & 84.70 & 50.41 & 10.79 \\
& & & OURS & 61.05 & \textbf{87.58} & \textbf{57.51} & \textbf{12.44} \\
\cmidrule(lr){3-8}

& & \multirow{2}{*}{MPDD} & Base & 53.06 & 75.52 & 29.07 & 5.33 \\
& & & OURS & 53.06 & \textbf{78.26} & \textbf{34.44} & \textbf{6.93} \\
\cmidrule(lr){3-8}

& & \multirow{2}{*}{BTAD} & Base & 79.97 & 82.60 & 44.86 & 12.27 \\
  & & & OURS & 79.97 & \textbf{86.80} & \textbf{48.28} & \textbf{22.13} \\
\bottomrule
\end{tabular}
}
\end{table*}

\subsubsection{AdaptCLIP}
\label{subsec:supp_adaptclip}

Table~\ref{tab:adaptclip_results} reports AdaptCLIP results.
The gains are more modest than in some FAPrompt or AnomalyCLIP settings, which is expected for a host that already includes stronger textual, visual, and prompt-query adaptation.
For ViT-L/14 OpenAI on BTAD, P-AUC/PRO/AP change from $93.80/65.41/41.15$ to $94.09/65.81/41.58$.
For ViT-L/14-336 on BTAD, they change from $93.84/70.22/41.79$ to $94.31/70.66/42.40$.
These are small but directionally consistent local-ranking gains, suggesting that a source-calibrated residual can still refine ambiguous responses without replacing the adapted host.
Other entries are mixed, especially when the baseline is already strong or unstable across seeds.
This behavior is consistent with the design of C-TED: it is intended to correct remaining hard-FP competition, not to override a well-calibrated detector or force large changes where little ambiguity remains.
We include AdaptCLIP as an important boundary case that clarifies the scope of the method.

\begin{table*}[h]
  \centering
  \caption{Results with \textbf{AdaptCLIP} as host.}
  \label{tab:adaptclip_results}
  \small
  \setlength{\tabcolsep}{4pt}
  \renewcommand{\arraystretch}{1.08}
  \resizebox{\textwidth}{!}{
  \begin{tabular}{lllc cccc}
  \toprule
  Host & Backbone & Target & Method & I-AUROC & P-AUROC & P-PRO & P-AP \\
  \midrule

  \multirow{32}{*}{\begin{tabular}[c]{@{}c@{}}AdaptCLIP\end{tabular}}

& \multirow{8}{*}{ViT-B/16+}
& \multirow{2}{*}{MVTec} & Base & 66.05$\pm$4.16 & \textbf{68.91$\pm$10.44} & \textbf{43.40$\pm$9.63} & 15.22$\pm$3.02 \\
& & & OURS & \textbf{66.06$\pm$4.14} & 68.63$\pm$10.58 & 42.75$\pm$9.76 & \textbf{15.41$\pm$3.01} \\
\cmidrule(lr){3-8}

& & \multirow{2}{*}{VisA} & Base
& \textbf{54.46$\pm$3.63} & \textbf{50.17$\pm$14.41} & \textbf{22.72$\pm$16.49} & \textbf{2.40$\pm$2.06} \\
& & & OURS
& 54.21$\pm$3.27 & 50.01$\pm$14.65 & 22.33$\pm$16.42 & 2.36$\pm$1.99 \\
\cmidrule(lr){3-8}

& & \multirow{2}{*}{MPDD} & Base & 57.00$\pm$3.18 & 69.97$\pm$7.55 & 35.12$\pm$5.41 & \textbf{6.10$\pm$2.76} \\
& & & OURS & \textbf{57.01$\pm$3.18} & \textbf{70.20$\pm$7.70} & \textbf{35.51$\pm$5.62} & 6.10$\pm$2.80 \\
\cmidrule(lr){3-8}

& & \multirow{2}{*}{BTAD} & Base & \textbf{62.52$\pm$12.53} & \textbf{61.03$\pm$12.90} & \textbf{25.20$\pm$11.36} & \textbf{8.05$\pm$7.55} \\
& & & OURS & 61.71$\pm$12.09 & 60.84$\pm$13.13 & 24.53$\pm$11.17 & 7.88$\pm$7.36 \\

  \cmidrule(lr){2-8}

& \multirow{8}{*}{ViT-L/14 OpenAI}
& \multirow{2}{*}{MVTec} & Base & 88.56$\pm$0.00 & 90.81$\pm$0.00 & 82.18$\pm$0.00 & 38.76$\pm$0.00 \\
& & & OURS & \textbf{89.01$\pm$0.00} & \textbf{90.96$\pm$0.00} & \textbf{82.73$\pm$0.00} & \textbf{39.61$\pm$0.00} \\
\cmidrule(lr){3-8}

& & \multirow{2}{*}{VisA} & Base
& \textbf{79.01$\pm$0.00} & 95.14$\pm$0.00 & 85.18$\pm$0.00 & 23.15$\pm$0.00 \\
& & & OURS
& 78.62$\pm$0.00 & \textbf{95.22$\pm$0.00} & \textbf{85.21$\pm$0.00} & \textbf{23.45$\pm$0.00} \\
\cmidrule(lr){3-8}

& & \multirow{2}{*}{MPDD} & Base & \textbf{70.49$\pm$0.00} & 95.33$\pm$0.00 & 85.95$\pm$0.00 & 19.94$\pm$0.00 \\
& & & OURS & 69.97$\pm$0.00 & \textbf{95.36$\pm$0.00} & \textbf{86.40$\pm$0.00} & \textbf{20.02$\pm$0.00} \\
\cmidrule(lr){3-8}

& & \multirow{2}{*}{BTAD} & Base & 85.72$\pm$0.00 & 93.80$\pm$0.00 & 65.41$\pm$0.00 & 41.15$\pm$0.00 \\
& & & OURS & \textbf{85.81$\pm$0.00} & \textbf{94.09$\pm$0.00} & \textbf{65.81$\pm$0.00} & \textbf{41.58$\pm$0.00} \\

  \cmidrule(lr){2-8}

  & \multirow{8}{*}{ViT-L/14-336}
& \multirow{2}{*}{MVTec} & Base & 93.48$\pm$0.00 & 90.93$\pm$0.00 & 83.54$\pm$0.00 & 38.30$\pm$0.00 \\
& & & OURS & \textbf{93.69$\pm$0.00} & \textbf{91.15$\pm$0.00} & \textbf{84.03$\pm$0.00} & \textbf{39.25$\pm$0.00} \\
  \cmidrule(lr){3-8}

  & & \multirow{2}{*}{VisA} & Base & \textbf{84.98}$\pm$0.00 & {95.67}$\pm$0.00 & 86.74$\pm$0.00 & {26.20} $\pm$0.00\\
  & & & OURS & {84.57} $\pm$0.00& \textbf{95.78}$\pm$0.00 & \textbf{86.97} $\pm$0.00& \textbf{26.67}$\pm$0.00 \\
  \cmidrule(lr){3-8}

  & & \multirow{2}{*}{MPDD} & Base & 73.57$\pm$0.00 & {95.94}$\pm$0.00 & 88.47 $\pm$0.00& 25.35$\pm$0.00 \\
  & & & OURS & \textbf{73.61}$\pm$0.00 & \textbf{95.98} $\pm$0.00& \textbf{88.52}$\pm$0.00 & \textbf{25.49}$\pm$0.00 \\
  \cmidrule(lr){3-8}

  & & \multirow{2}{*}{BTAD} & Base & \textbf{91.04}$\pm$0.00 & {93.84}$\pm$0.00 & {70.22} $\pm$0.00& {41.79}$\pm$0.00 \\
  & & & OURS & 90.78$\pm$0.00 & \textbf{94.31}$\pm$0.00 & \textbf{70.66}$\pm$0.00 & \textbf{42.40}$\pm$0.00 \\

  \cmidrule(lr){2-8}

& \multirow{8}{*}{ViT-H/14}
& \multirow{2}{*}{MVTec} & Base & 69.23$\pm$4.92 & 72.62$\pm$5.89 & 49.30$\pm$8.74 & 15.64$\pm$4.24 \\
& & & OURS & \textbf{69.24$\pm$4.99} & \textbf{72.70$\pm$6.20} & \textbf{49.34$\pm$9.14} & \textbf{16.14$\pm$4.50} \\
\cmidrule(lr){3-8}

& & \multirow{2}{*}{VisA} & Base
& \textbf{63.39$\pm$2.49} & \textbf{78.98$\pm$7.84} & \textbf{55.02$\pm$9.92} & 14.92$\pm$2.67 \\
& & & OURS
& 63.26$\pm$2.57 & 78.70$\pm$8.20 & 54.75$\pm$9.98 & \textbf{15.64$\pm$2.68} \\
\cmidrule(lr){3-8}

& & \multirow{2}{*}{MPDD} & Base & \textbf{55.43$\pm$5.11} & \textbf{79.76$\pm$7.72} & 47.91$\pm$15.70 & 9.55$\pm$3.77 \\
& & & OURS & 55.32$\pm$4.92 & 79.64$\pm$7.97 & \textbf{48.19$\pm$16.07} & \textbf{9.70$\pm$3.79} \\
\cmidrule(lr){3-8}

& & \multirow{2}{*}{BTAD} & Base & \textbf{55.73$\pm$8.33} & 72.71$\pm$3.65 & 35.63$\pm$6.81 & 11.26$\pm$6.63 \\
& & & OURS & 55.46$\pm$7.92 & \textbf{73.28$\pm$3.82} & \textbf{36.06$\pm$6.86} & \textbf{11.57$\pm$6.83} \\

  \bottomrule
  \end{tabular}
  }
  \end{table*}

\subsubsection{AdaCLIP}
\label{subsec:supp_adaclip}

Table~\ref{tab:adaclip_results} reports AdaCLIP results.
Several settings show useful PRO gains even when P-AUC or AP changes are mixed.
For ViT-L/14 OpenAI on MVTec, PRO improves from $44.32$ to $49.01$.
For ViT-L/14-336 on BTAD, PRO improves from $19.26$ to $27.81$.
These gains indicate that source-conditioned correction can improve region-level localization quality in some AdaCLIP settings, even when threshold-free pixel ranking or sparse-pixel AP does not improve uniformly.
This distinction is important because AdaCLIP already uses image-conditioned prompting, so the remaining errors may differ from those of simpler prompt-tuned hosts.
In such cases, C-TED may help mainly by reducing region-level hard-FP competition rather than by globally reshaping the entire pixel-score distribution.
However, some P-AUC and AP entries remain mixed, especially on weaker or already saturated configurations.
This supports a bounded interpretation: C-TED is a local evidence-ranking correction whose benefit depends on the amount and structure of remaining hard-FP competition.
We therefore use AdaCLIP as an intermediate case between strong positive hosts and boundary hosts, showing that the correction can help but should not be read as uniformly improving every metric.

\begin{table*}[h]
\centering
\caption{Results with \textbf{AdaCLIP} as host.}
\label{tab:adaclip_results}
\small
\setlength{\tabcolsep}{4pt}
\renewcommand{\arraystretch}{1.08}
\resizebox{\textwidth}{!}{
\begin{tabular}{lllc cccc}
\toprule
Host & Backbone & Target & Method & I-AUROC & P-AUROC & P-PRO & P-AP \\
\midrule

\multirow{32}{*}{\begin{tabular}[c]{@{}c@{}}AdaCLIP\end{tabular}}

& \multirow{8}{*}{ViT-B/16+}
& \multirow{2}{*}{MVTec} & Base & 48.53$\pm$7.16 & \textbf{50.32$\pm$6.37} & 19.74$\pm$5.72 & 4.27$\pm$0.97 \\
& & & OURS & \textbf{49.67$\pm$8.24} & 50.26$\pm$6.45 & \textbf{20.81$\pm$5.70} & \textbf{4.46$\pm$1.17} \\
\cmidrule(lr){3-8}

& & \multirow{2}{*}{VisA} & Base & 49.19$\pm$1.25 & 57.15$\pm$7.57 & 25.60$\pm$6.46 & 1.16$\pm$0.35 \\
& & & OURS & \textbf{50.83$\pm$1.63} & \textbf{58.70$\pm$6.52} & \textbf{28.22$\pm$5.92} & \textbf{1.32$\pm$0.39} \\
\cmidrule(lr){3-8}

& & \multirow{2}{*}{MPDD} & Base & \textbf{43.82$\pm$2.60} & 42.46$\pm$10.95 & 14.05$\pm$6.18 & \textbf{2.82$\pm$0.87} \\
& & & OURS & 42.64$\pm$0.32 & \textbf{42.57$\pm$9.39} & \textbf{15.00$\pm$5.11} & 2.61$\pm$0.71 \\
\cmidrule(lr){3-8}

& & \multirow{2}{*}{BTAD} & Base & 64.80$\pm$9.50 & 51.09$\pm$4.34 & 14.84$\pm$3.01 & 3.58$\pm$0.78 \\
& & & OURS & \textbf{66.98$\pm$10.31} & \textbf{52.24$\pm$4.57} & \textbf{15.67$\pm$2.84} & \textbf{3.68$\pm$0.64} \\

\cmidrule(lr){2-8}

& \multirow{8}{*}{ViT-L/14 OpenAI}
& \multirow{2}{*}{MVTec} & Base & 84.66 & 89.54 & 44.32 & 39.36 \\
& & & OURS & \textbf{87.89} & \textbf{89.55} & \textbf{49.01} & \textbf{39.51} \\
& & \multirow{2}{*}{VisA} & Base
& 79.47$\pm$0.13 & \textbf{95.57$\pm$0.00} & 54.38$\pm$0.00 & \textbf{26.45$\pm$0.00} \\
& & & OURS
& \textbf{80.70$\pm$0.06} & 95.49$\pm$0.00 & \textbf{66.02$\pm$0.55} & 25.63$\pm$0.01 \\
\cmidrule(lr){3-8}

& & \multirow{2}{*}{MPDD} & Base
& 67.38$\pm$0.11 & 95.91$\pm$0.00 & 31.36$\pm$0.00 & \textbf{25.44$\pm$0.00} \\
& & & OURS
& \textbf{67.67$\pm$0.22} & \textbf{95.98$\pm$0.00} & \textbf{46.51$\pm$0.49} & 25.26$\pm$0.00 \\
\cmidrule(lr){3-8}

& & \multirow{2}{*}{BTAD} & Base
& \textbf{87.14$\pm$0.14} & \textbf{91.88$\pm$0.00} & 9.70$\pm$0.00 & \textbf{42.06$\pm$0.00} \\
& & & OURS
& 86.19$\pm$0.07 & 91.51$\pm$0.04 & \textbf{14.04$\pm$0.31} & 40.51$\pm$0.04 \\
\cmidrule(lr){2-8}

& \multirow{8}{*}{ViT-L/14-336}
& \multirow{2}{*}{MVTec} & Base & 89.93$\pm$0.06 & \textbf{89.89$\pm$0.00} & 44.03$\pm$0.00 & 41.60$\pm$0.00 \\
& & & OURS & \textbf{91.63$\pm$0.01} & 89.85$\pm$0.00 & \textbf{47.86$\pm$0.08} & \textbf{41.84$\pm$0.01} \\
\cmidrule(lr){3-8}

& & \multirow{2}{*}{VisA} & Base & 85.99$\pm$0.25 & \textbf{95.82$\pm$0.00} & 54.13$\pm$0.00 & \textbf{31.48$\pm$0.00} \\
& & & OURS & \textbf{86.38$\pm$0.02} & 95.72$\pm$0.00 & \textbf{65.61$\pm$0.84} & 31.31$\pm$0.01 \\
\cmidrule(lr){3-8}

& & \multirow{2}{*}{MPDD} & Base & 68.96$\pm$0.23 & 96.09$\pm$0.00 & 30.48$\pm$0.00 & 29.76$\pm$0.00 \\
& & & OURS & \textbf{69.41$\pm$0.21} & \textbf{96.33$\pm$0.00} & \textbf{44.55$\pm$0.83} & \textbf{29.92$\pm$0.01} \\
\cmidrule(lr){3-8}

& & \multirow{2}{*}{BTAD} & Base & \textbf{90.41$\pm$0.25} & \textbf{93.66$\pm$0.00} & 19.26$\pm$0.00 & \textbf{48.20$\pm$0.00} \\
& & & OURS & 90.04$\pm$0.20 & 93.49$\pm$0.03 & \textbf{27.81$\pm$0.67} & 46.85$\pm$0.02 \\

\cmidrule(lr){2-8}

& \multirow{8}{*}{ViT-H/14}
& \multirow{2}{*}{MVTec} & Base & 47.24$\pm$0.48 & \textbf{51.14$\pm$3.01} & \textbf{19.80$\pm$3.19} & \textbf{4.76$\pm$0.33} \\
& & & OURS & \textbf{47.46$\pm$0.64} & 49.25$\pm$1.91 & 18.97$\pm$1.15 & 4.74$\pm$0.04 \\
\cmidrule(lr){3-8}

& & \multirow{2}{*}{VisA} & Base & 46.32$\pm$4.34 & 53.92$\pm$2.35 & 21.38$\pm$4.10 & 1.16$\pm$0.28 \\
& & & OURS & \textbf{48.18$\pm$3.71} & \textbf{55.24$\pm$2.83} & \textbf{24.41$\pm$4.36} & \textbf{1.31$\pm$0.26} \\
\cmidrule(lr){3-8}

& & \multirow{2}{*}{MPDD} & Base & 38.37$\pm$5.28 & 49.09$\pm$9.25 & 17.95$\pm$8.72 & 2.73$\pm$1.57 \\
& & & OURS & \textbf{40.06$\pm$3.87} & \textbf{52.16$\pm$8.18} & \textbf{20.96$\pm$6.24} & \textbf{2.93$\pm$1.44} \\
\cmidrule(lr){3-8}

& & \multirow{2}{*}{BTAD} & Base & 62.93$\pm$13.69 & 57.21$\pm$5.35 & 23.13$\pm$3.62 & 4.29$\pm$1.19 \\
& & & OURS & \textbf{64.86$\pm$14.71} & \textbf{57.59$\pm$4.83} & \textbf{25.63$\pm$4.88} & \textbf{4.71$\pm$1.18} \\

\bottomrule
\end{tabular}
}
\end{table*}

\subsubsection{BayesPFL}
\label{subsec:supp_bayespfl}

Table~\ref{tab:bayespfl_results} reports BayesPFL results.
Compared with the other hosts, BayesPFL shows more mixed pixel-level behavior.
Several settings show image-level gains: for example, ViT-L/14-336 on MVTec improves I-AUC from $92.69$ to $96.40$, and ViT-L/14-336 on BTAD improves I-AUC from $77.89$ to $89.92$.
Pixel-level changes are often smaller or mixed.
For instance, ViT-L/14 OpenAI on BTAD has a small PRO gain from $64.28$ to $65.57$, while AP decreases from $35.98$ to $34.30$.
This table is useful because it documents a boundary regime: when a host already provides relatively stable local ranking or when the remaining error is not primarily hard-FP competition, C-TED may yield limited or mixed pixel-level changes.
This behavior is consistent with our failure-conditioned interpretation, since the residual correction is expected to help most when hard-normal evidence remains close to true defects.
It also shows why we report image-level and pixel-level metrics separately: improvements in image-level ranking do not necessarily imply cleaner dense localization.
Conversely, small or mixed pixel-level changes do not by themselves invalidate the evidence-decoding view, but indicate that the measured failure may be weaker or different in that host.
BayesPFL therefore helps define the practical scope of C-TED rather than serving as a primary positive example.
We therefore use BayesPFL to clarify scope rather than to overstate uniform improvement.

\begin{table*}[h]
\centering
\caption{Results with \textbf{BayesPFL} as host using calibrated TED.}
\label{tab:bayespfl_results}
\small
\setlength{\tabcolsep}{4pt}
\renewcommand{\arraystretch}{1.08}
\resizebox{\textwidth}{!}{
\begin{tabular}{lllc cccc}
\toprule
Host & Backbone & Target & Method & I-AUROC & P-AUROC & P-PRO & P-AP \\
\midrule

\multirow{32}{*}{\begin{tabular}[c]{@{}c@{}}BayesPFL\end{tabular}}

& \multirow{8}{*}{ViT-B/16+}
& \multirow{2}{*}{MVTec} & Base & 49.68$\pm$1.01 & \textbf{49.73$\pm$3.18} & \textbf{17.14$\pm$3.94} & \textbf{4.29$\pm$0.80} \\
& & & OURS & \textbf{52.49$\pm$2.56} & 49.67$\pm$3.19 & 17.00$\pm$3.86 & 4.26$\pm$0.78 \\
\cmidrule(lr){3-8}

& & \multirow{2}{*}{VisA} & Base & \textbf{51.41$\pm$1.85} & \textbf{54.71$\pm$3.85} & \textbf{18.93$\pm$5.43} & \textbf{0.90$\pm$0.26} \\
& & & OURS & 50.20$\pm$2.31 & 54.60$\pm$3.67 & 18.87$\pm$5.32 & 0.88$\pm$0.24 \\
\cmidrule(lr){3-8}

& & \multirow{2}{*}{MPDD} & Base & 50.35$\pm$4.02 & \textbf{58.47$\pm$9.08} & \textbf{22.52$\pm$9.48} & \textbf{2.39$\pm$0.29} \\
& & & OURS & \textbf{51.26$\pm$4.88} & 58.27$\pm$8.73 & 22.43$\pm$9.22 & 2.38$\pm$0.27 \\
\cmidrule(lr){3-8}

& & \multirow{2}{*}{BTAD} & Base & \textbf{51.46$\pm$5.29} & \textbf{43.32$\pm$0.34} & \textbf{9.79$\pm$0.76} & 2.86$\pm$0.78 \\
& & & OURS & 49.85$\pm$7.65 & 43.29$\pm$0.31 & 9.75$\pm$0.74 & \textbf{2.87$\pm$0.79} \\

\cmidrule(lr){2-8}

& \multirow{8}{*}{ViT-L/14 OpenAI}
& \multirow{2}{*}{MVTec} & Base & 90.42$\pm$0.16 & 97.28$\pm$0.03 & \textbf{90.20$\pm$0.05} & \textbf{57.51$\pm$0.11} \\
& & & OURS & \textbf{94.82$\pm$0.08} & \textbf{97.28$\pm$0.03} & 90.14$\pm$0.05 & 57.16$\pm$0.08 \\
\cmidrule(lr){3-8}

& & \multirow{2}{*}{VisA} & Base & 75.89$\pm$0.62 & \textbf{94.53$\pm$0.08} & \textbf{82.14$\pm$0.07} & \textbf{18.99$\pm$0.24} \\
& & & OURS & \textbf{75.90$\pm$0.28} & 94.51$\pm$0.08 & 82.06$\pm$0.06 & 18.80$\pm$0.22 \\
\cmidrule(lr){3-8}

& & \multirow{2}{*}{MPDD} & Base & 70.07$\pm$0.23 & \textbf{95.73$\pm$0.06} & 86.24$\pm$0.14 & \textbf{24.92$\pm$0.32} \\
& & & OURS & \textbf{70.65$\pm$0.91} & 95.72$\pm$0.06 & \textbf{86.26$\pm$0.21} & 24.81$\pm$0.31 \\
\cmidrule(lr){3-8}

& & \multirow{2}{*}{BTAD} & Base & 73.71$\pm$2.08 & \textbf{87.94$\pm$0.27} & 64.28$\pm$0.31 & \textbf{35.98$\pm$0.28} \\
& & & OURS & \textbf{83.90$\pm$0.80} & 87.89$\pm$0.28 & \textbf{64.33$\pm$0.30} & 35.62$\pm$0.28 \\

\cmidrule(lr){2-8}

& \multirow{8}{*}{ViT-L/14-336}
& \multirow{2}{*}{MVTec} & Base & 92.69$\pm$0.32 & 97.35$\pm$0.06 & 92.34$\pm$0.01 & 64.22$\pm$0.28 \\
& & & OURS & \textbf{96.40$\pm$0.11} & \textbf{97.36$\pm$0.06} & \textbf{92.35$\pm$0.02} & \textbf{64.24$\pm$0.27} \\
\cmidrule(lr){3-8}

& & \multirow{2}{*}{VisA} & Base & 79.51$\pm$0.24 & 95.51$\pm$0.02 & 89.25$\pm$0.06 & \textbf{25.30$\pm$0.17} \\
& & & OURS & \textbf{80.74$\pm$0.51} & \textbf{95.52$\pm$0.02} & \textbf{89.25$\pm$0.06} & 25.19$\pm$0.18 \\
\cmidrule(lr){3-8}

& & \multirow{2}{*}{MPDD} & Base & \textbf{78.32$\pm$1.11} & \textbf{96.89$\pm$0.02} & \textbf{91.27$\pm$0.09} & \textbf{29.99$\pm$0.18} \\
& & & OURS & 74.51$\pm$0.30 & 96.88$\pm$0.02 & 91.25$\pm$0.16 & 29.97$\pm$0.19 \\
\cmidrule(lr){3-8}

& & \multirow{2}{*}{BTAD} & Base & 77.89$\pm$0.88 & 86.98$\pm$0.24 & \textbf{71.61$\pm$0.15} & \textbf{35.05$\pm$0.61} \\
& & & OURS & \textbf{89.92$\pm$1.14} & \textbf{86.98$\pm$0.26} & 71.56$\pm$0.10 & 34.91$\pm$0.65 \\

\cmidrule(lr){2-8}

& \multirow{8}{*}{ViT-H/14}
& \multirow{2}{*}{MVTec} & Base & 46.82$\pm$4.08 & 51.74$\pm$9.31 & 17.43$\pm$7.07 & 3.86$\pm$1.50 \\
& & & OURS & \textbf{53.02$\pm$0.59} & \textbf{51.83$\pm$9.36} & \textbf{17.55$\pm$7.09} & \textbf{3.87$\pm$1.52} \\
\cmidrule(lr){3-8}

& & \multirow{2}{*}{VisA} & Base & 49.62$\pm$1.15 & 47.76$\pm$11.74 & 13.72$\pm$6.40 & 0.73$\pm$0.38 \\
& & & OURS & \textbf{50.19$\pm$0.90} & \textbf{47.80$\pm$11.81} & \textbf{13.77$\pm$6.46} & \textbf{0.74$\pm$0.39} \\
\cmidrule(lr){3-8}

& & \multirow{2}{*}{MPDD} & Base & \textbf{49.60$\pm$2.81} & 42.50$\pm$7.10 & 10.58$\pm$4.90 & 2.04$\pm$0.50 \\
& & & OURS & 49.51$\pm$7.04 & \textbf{42.56$\pm$7.26} & \textbf{10.66$\pm$4.97} & \textbf{2.04$\pm$0.50} \\
\cmidrule(lr){3-8}

& & \multirow{2}{*}{BTAD} & Base & 47.20$\pm$6.28 & 48.56$\pm$4.20 & \textbf{12.29$\pm$5.17} & \textbf{2.84$\pm$0.19} \\
& & & OURS & \textbf{50.70$\pm$4.26} & \textbf{48.59$\pm$4.17} & 12.22$\pm$4.90 & 2.84$\pm$0.19 \\

\bottomrule
\end{tabular}
}
\end{table*}

\subsection{Full Recipe-Retuning Results}
\label{subsec:supp_layer_retrain}

Table~\ref{tab:layer_recipe_retrain} reports the full bidirectional recipe-retuning diagnostic.
Each cell reports I-AUC / P-AUC / PRO / AP.
The table complements the main-paper recipe summary by showing that C-TED can remain useful after the host is retrained under changed layer recipes and source banks are rebuilt for the resulting representation.
This setting is stricter than a fixed-host swap because the host is allowed to adapt to the new recipe before C-TED is applied.
Therefore, the remaining gains are less likely to be explained solely by a mismatched feature hook or an unfavorable layer choice.
For FAPrompt, C-TED improves AP across all listed layer recipes in both directions.
For example, in the VisA$\rightarrow$MVTec direction, Full(B) AP improves from $22.35$ to $30.94$.
In the MVTec$\rightarrow$VisA direction, Full(B) AP improves from $13.61$ to $19.23$.
For AdaCLIP, several recipes show clear PRO gains: VisA$\rightarrow$MVTec L-r improves PRO from $34.83$ to $46.27$, and MVTec$\rightarrow$VisA Full(B) improves PRO from $54.13$ to $65.18$.
The gains appear in different metrics for different hosts: FAPrompt shows particularly clear AP improvements, while AdaCLIP often benefits more in PRO.
This suggests that C-TED does not simply optimize one metric-specific artifact, but can improve different aspects of local ranking depending on the host.
At the same time, the effect is not uniform across all cells, so recipe retuning remains an important factor in the host's baseline behavior.
These results support the main claim that representation selection and evidence decoding are distinct.
Changing or retraining the layer recipe can shift host responses, but it does not necessarily remove hard-FP ranking errors.
C-TED can still improve local ranking after recipe retuning, although the magnitude remains host-, recipe-, and metric-dependent.

\begin{table*}[t]
\centering
\caption{
Bidirectional layer-recipe retraining diagnostic.
Each cell reports I-AUC / P-AUC / PRO / AP.
B, T, and C denote the retrained host baseline, train-free TED, and source-calibrated TED, respectively.
Full(B) is the official full-layer baseline recipe; EL-r, M2-r, L2-r, and L-r denote checkpoints retrained with early+last, middle-2, last-2, and last-layer recipes.
}
\label{tab:layer_recipe_retrain}
\small
\setlength{\tabcolsep}{2.2pt}
\renewcommand{\arraystretch}{1.03}
\resizebox{\textwidth}{!}{
\begin{tabular}{lll|c|c|c}
\toprule
Transfer & Host & Recipe & B & T & C \\
\midrule
\multirow{10}{*}{VisA$\rightarrow$MVTec}
& \multirow{5}{*}{AdaCLIP}
& Full(B)
& 89.93 / \textbf{89.89} / 44.03 / 41.60
& 91.43 / 89.88 / 43.60 / 41.46
& \textbf{91.64} / 89.86 / \textbf{47.97} / \textbf{41.85} \\
& & EL-r
& 87.83 / 80.21 / 18.18 / 22.40
& 88.74 / 80.21 / 17.94 / 22.39
& \textbf{88.84} / \textbf{80.96} / \textbf{26.25} / \textbf{23.81} \\
& & M2-r
& \textbf{89.26} / 87.62 / 23.50 / 36.18
& 87.74 / 87.61 / 23.63 / 36.12
& 87.74 / \textbf{87.79} / \textbf{25.42} / \textbf{36.40} \\
& & L2-r
& \textbf{90.58} / 89.47 / 15.19 / 36.35
& 89.07 / 89.47 / 15.14 / 36.31
& 88.69 / \textbf{90.84} / \textbf{22.78} / \textbf{39.28} \\
& & L-r
& 90.40 / 88.96 / 34.83 / 35.78
& \textbf{91.99} / 88.96 / 34.66 / 35.77
& 91.79 / \textbf{89.15} / \textbf{46.27} / \textbf{36.77} \\
\cmidrule(lr){2-6}
& \multirow{5}{*}{FAPrompt}
& Full(B)
& \textbf{91.45} / 89.95 / 73.57 / 22.35
& \textbf{91.45} / 40.41 / 25.32 / 8.16
& \textbf{91.45} / \textbf{90.19} / \textbf{76.83} / \textbf{30.94} \\
& & EL-r
& \textbf{91.40} / 90.49 / 74.70 / 23.58
& \textbf{91.40} / 42.46 / 26.23 / 9.25
& \textbf{91.40} / \textbf{90.78} / \textbf{77.16} / \textbf{32.40} \\
& & M2-r
& \textbf{92.19} / 90.05 / 74.35 / 22.58
& \textbf{92.19} / 41.66 / 25.87 / 8.79
& \textbf{92.19} / \textbf{90.52} / \textbf{76.54} / \textbf{31.46} \\
& & L2-r
& \textbf{92.61} / 90.08 / 72.73 / 22.33
& \textbf{92.61} / 45.33 / 30.26 / 10.78
& \textbf{92.61} / \textbf{90.61} / \textbf{76.63} / \textbf{31.07} \\
& & L-r
& \textbf{90.93} / 89.63 / 74.01 / 23.01
& \textbf{90.93} / 44.35 / 29.27 / 10.40
& \textbf{90.93} / \textbf{89.93} / \textbf{76.08} / \textbf{31.68} \\
\midrule
\multirow{10}{*}{MVTec$\rightarrow$VisA}
& \multirow{5}{*}{AdaCLIP}
& Full(B)
& 85.65 / \textbf{95.82} / 54.13 / 31.48
& \textbf{87.00} / \textbf{95.82} / 54.43 / \textbf{31.54}
& 86.36 / 95.72 / \textbf{65.18} / 31.32 \\
& & EL-r
& 82.05 / 90.56 / 57.48 / 20.45
& \textbf{83.10} / \textbf{91.38} / \textbf{59.64} / \textbf{20.79}
& 82.92 / 91.33 / 57.65 / 20.63 \\
& & M2-r
& 78.89 / \textbf{95.17} / 58.12 / 29.03
& 81.56 / \textbf{95.17} / 58.10 / 29.03
& \textbf{81.90} / 95.16 / \textbf{60.36} / \textbf{29.11} \\
& & L2-r
& 73.87 / 92.36 / 63.13 / 22.25
& \textbf{78.12} / 92.36 / 63.16 / 22.26
& 77.80 / \textbf{93.76} / \textbf{68.26} / \textbf{22.61} \\
& & L-r
& 81.47 / 92.37 / 50.61 / 21.52
& \textbf{83.08} / 92.37 / 50.44 / \textbf{21.56}
& 81.56 / \textbf{93.32} / \textbf{58.91} / 21.05 \\
\cmidrule(lr){2-6}
& \multirow{5}{*}{FAPrompt}
& Full(B)
& \textbf{82.23} / 95.33 / 79.34 / 13.61
& \textbf{82.23} / 93.46 / 78.07 / 13.52
& \textbf{82.23} / \textbf{95.60} / \textbf{82.08} / \textbf{19.23} \\
& & EL-r
& \textbf{83.25} / 95.36 / 79.45 / 13.15
& \textbf{83.25} / 92.14 / 78.08 / 16.06
& \textbf{83.25} / \textbf{95.59} / \textbf{80.62} / \textbf{18.45} \\
& & M2-r
& \textbf{84.34} / 95.36 / 78.13 / 13.37
& \textbf{84.34} / 90.41 / 75.46 / 15.77
& \textbf{84.34} / \textbf{95.64} / \textbf{81.54} / \textbf{18.57} \\
& & L2-r
& \textbf{84.40} / 95.07 / 78.82 / 13.10
& \textbf{84.40} / 93.22 / 78.98 / 14.72
& \textbf{84.40} / \textbf{95.39} / \textbf{82.87} / \textbf{18.66} \\
& & L-r
& \textbf{82.30} / 95.32 / 77.29 / 13.46
& \textbf{82.30} / 92.68 / 79.20 / 15.81
& \textbf{82.30} / \textbf{95.53} / \textbf{81.29} / \textbf{18.71} \\
\bottomrule
\end{tabular}
}
\end{table*}

\section{Limitations and Broader Impacts}
\label{sec:supp_limitations_impacts}

\paragraph{Limitations.}
\textsc{TED} is designed for local anomaly maps, not universal image-level screening.
The method preserves the host detector and modifies only the local evidence readout; therefore, image-level performance can still depend on the host's global branch, pooling rule, or score aggregation strategy.
Although we report I-AUROC for transparency, the primary claim of \textsc{TED} is pixel-level localization under source-to-target transfer.

\textsc{TED} also depends on source defect and hard-FP banks.
Its correction quality can depend on whether the source split contains defect evidence and hard-normal patterns that are relevant to the target domain.
If the target domain contains defect types or normal structures not represented in the source banks, the support comparison may become less reliable.
Similarly, if the source anomaly masks are noisy or the mined hard-FP patches do not reflect the dominant target false positives, the residual correction may provide limited benefit.
Our fixed-bank and source-supervision controls reduce some confounding explanations, but they do not eliminate the need for representative source evidence.

The method is also evaluated within the VLM backbones and CLIP-AD hosts considered in this work.
\textsc{TED} uses a weaker interface than many prompt-learning pipelines, requiring patch-level visual features and a normal-versus-anomaly text response, but this should not be read as universal backbone-agnostic anomaly detection.
Backbones without meaningful patch-level features, unstable text responses, or incompatible feature scales may require additional validation.
Finally, the source-calibrated residual introduces extra source-side calibration and bank storage, even though the retained-bank sensitivity experiment suggests that moderate bank sizes are sufficient in our tested settings.
Future work should study source-bank construction, image-level aggregation, broader host families, failure cases, and deployment-time monitoring under larger distribution shifts.

\paragraph{Broader impacts.}
\textsc{TED} is intended for industrial anomaly localization and may reduce false-positive burden by improving local defect ranking under domain shift.
In practice, it can help human inspectors focus on suspicious regions and provide more interpretable local evidence, especially when retraining large host models is costly.
At the same time, over-reliance on automated localization can be harmful: missed defects may affect product quality or safety, and false alarms may increase inspection cost or reduce operator trust.
These risks are higher when source defect and hard-FP banks do not cover the target inspection conditions, including new materials, lighting, imaging artifacts, or defect types.
\textsc{TED} should therefore be used as decision support with human oversight, periodic target-domain validation, and monitoring of false positives and missed defects.
The work uses public industrial inspection datasets and does not involve personal, biometric, medical, or otherwise sensitive data; the main concern is reliability in quality- or safety-critical inspection workflows.


\end{document}